\documentclass[preprint,12pt]{elsarticle}

\usepackage{amssymb}
\usepackage{amsmath}

\usepackage{cite}
\usepackage{hyperref}
\usepackage{amsthm, amssymb, amsmath}
\usepackage{graphicx}
\usepackage{epstopdf}
\usepackage{float}
\usepackage{rotating}
\usepackage{subfigure}
\usepackage[lined,ruled,linesnumbered,commentsnumbered]{algorithm2e}
\usepackage{cleveref}
\usepackage{booktabs}
\usepackage{amsmath,breqn}
\usepackage{multirow}
\usepackage{amsmath}
\usepackage{pdflscape}
\usepackage{color, colortbl}
\usepackage{threeparttable} 
\usepackage{array}

\begin{document}

\begin{frontmatter}



\title{CR-BLEA: Contrastive Ranking for Adaptive Resource Allocation in Bilevel Evolutionary Algorithms}





\author[1]{Dejun Xu}

\author[1]{Jijia Chen}
\address[1]{School of Informatics, Xiamen University, Xiamen, 361005, China.}

\author[2]{Gary G. Yen}
\address[2]{Department of Artificial Intelligence, Sichuan University, Chengdu, 6100065, China}

\author[1]{Min Jiang\corref{cor1}}
\ead{minjiang@xmu.edu.cn}

\cortext[cor1]{Corresponding author.}


\begin{abstract}
Bilevel optimization poses a significant computational challenge due to its nested structure, where each upper-level candidate solution requires solving a corresponding lower-level problem. While evolutionary algorithms (EAs) are effective at navigating such complex landscapes, their high resource demands remain a key bottleneck—particularly the redundant evaluation of numerous unpromising lower-level tasks. Despite recent advances in multitasking and transfer learning, resource waste persists. To address this issue, we propose a novel resource allocation framework for bilevel EAs that selectively identifies and focuses on promising lower-level tasks. Central to our approach is a contrastive ranking network that learns relational patterns between paired upper- and lower-level solutions online. This knowledge guides a reference-based ranking strategy that prioritizes tasks for optimization and adaptively controls resampling based on estimated population quality. Comprehensive experiments across five state-of-the-art bilevel algorithms show that our framework significantly reduces computational cost while preserving—or even enhancing—solution accuracy. This work offers a generalizable strategy to improve the efficiency of bilevel EAs, paving the way for more scalable bilevel optimization.
\end{abstract}








\end{frontmatter}



\section{Introduction}
Bilvel optimization problems (BLOPs) are characterized by their hierarchical structure and interactive nature.
Within the bilevel framework, the lower-level optimization problem is affected by the upper-level decision, and its optimization results, in turn, affect the decision of the upper level.
Therefore, the upper level needs to seek the optimal solution while considering the response from the lower level.

BLOPs are prevalent in real-world applications, such as  logistics management involving depots location selection and distribution route planning \citep{ghasemi2023bi}, and adversarially robust training for machine learning models which involves defensive capability optimization and attack perturbation generation \citep{zhang2024introduction,zhang2022revisiting}.
Taking neural architecture search (NAS) as an example, the goal is to identify an optimal network architecture that maximizes model performance under given constraints such as computational resources.
The upper level of the search process focuses on selecting the structural parameters, including the number, types and connections of layers.
Given the architecture defined by the upper level, the lower level then optimizes the training loss of the model by adjusting its weights and bias.
Based on the refined model provided by the lower level, the upper level iteratively evaluates and adjusts the architecture parameters to further improve model performance.
The interactive  nature of BLOPs renders them significantly more complex than single-level optimization problems and has been proven to be NP-hard \citep{bard1991some}.

The complexity of BLOPs falls well within the capabilities of evolutionary algorithms (EAs), which possess global search capability that enables them to search for satisfactory solutions of various BLOPs. 
However, when handling BLOPs, bilevel evolutionary algorithms (BLEAs) require substantial computational resources embodied as the number of function evaluations (FEs).
On one hand, this is due to the inherent requirement of evolutionary algorithms for extensive iterations even when addressing non-hierarchical problems \citep{camacho2023metaheuristics}, let alone bilevel ones.
Moreover, to search for valid optimal solutions, evolutionary process needs to be executed cyclically at both the upper and lower levels. 
Notably, the number of lower-level iterations significantly surpasses that of upper-level iterations, as each individual in the upper-level population generates an optimization problem that requires multiple iterations to search for the optimal lower-level solution.

The substantial resource consumption has seriously limited the application of evolutionary algorithms in real-world bilevel optimization problems, and also makes reducing function evaluation consumption in bilevel optimization a recent research focus.
Various strategies such as co-evolution and knowledge transfer have been introduced to improve bilevel optimization over the past few years \citep{chaabani2023solving}.
For instance, a search distribution sharing mechanism is implemented between the upper and lower levels, which promotes the lower-level CMA-ES optimizer with a priori knowledge \citep{he2018evolutionary}.
In  \citep{chen2021transfer} and \citep{chen2023evolutionary}, multiple lower-level search processes are parallelized, where transfer learning and multi-objective transformation are performed to improve search efficiency.
Additionally, multitasking has been carried out to facilitate collaboration among lower-level tasks \citep{huang2023bilevel,gupta2015evolutionary}.

To some extent, these methods have facilitated the lower-level optimization processes that are large in scale, thereby reducing the resource consumption of bilevel evolutionary algorithms.
However, there is a ceiling to these improvements.
Essentially, bilevel evolutionary optimization is a veiled process with an implicit selection mechanism.
Based on the implicit constraints defined in BLOPs, in order to determine which individuals should be eliminated from an upper-level population, BLEA needs to optimize each lower-level problem $f(x_u,x_l)$ derived from the upper-level individuals, and find the corresponding lower-level optimal solution $x_l^*$ for each $x_u$.
Only with the corresponding $x_l^*$ can an upper-level individual $x_u$ be subsequently evaluated for its upper-level fitness by $F(x_u,x_l^*)$, and be determined whether to be retained.
Obviously, this differs from the explicit selection mechanism in single-level evolution as considerable effort must be invested to compute $x_l^*$ before evaluating $x_u$.

In fact, once the structure of the lower-level problem is determined, $x_l^*$, as the optimal solution of the lower-level problem derived from the upper-level individual, is essentially a function of $x_u$, but it needs to be revealed through the lower-level search.
Recent attempts to reduce resource consumption mentioned above are mainly focused on facilitating the process of revealing $x_l^*$ using various knowledge transfer strategies\citep{chen2021transfer,chen2023evolutionary,huang2023bilevel}.
However, it is disappointing that some efforts are inevitably wasted, as typically a large proportion of $x_u$ are discarded in the subsequent upper-level environmental selection involving $F(x_u,x_l^*)$.
Although the efficiency of the process of revealing $x_l^*$ is improved, the resources expended in revealing $x_l^*$ for those discarded $x_u$ remain on a massive scale.

Consequently, the process of revealing $x_l^*$ is considered in this work as the ``veil” of bilevel evolutionary optimization, which leads to futile efforts on some unpromising lower-level tasks and ultimately results in a waste of resources.
To break the efficiency ceiling of BLEAs as discussed above, this veil needs to be removed.
In this work, 
we propose a contrastive ranking-based framework (CR-BLEA) that can be integrated within various bilevel optimization strategies to reduce their computational cost. 
CR-BLEA leverages the knowledge generated during the evolutionary process to identify and allocate computational resources to promising tasks.
Generally, the contributions of this work can be summarized as follows:

$\bullet$ A knowledge-driven resource allocation framework is proposed, in which only the promising lower-level tasks are selected for execution. 
The framework significantly reduces the substantial resource consumption in bilevel evolutionary optimization and allows the integration of any nested BLEA, including recent approaches based on knowledge transfer.

$\bullet$ A contrastive neural network is designed to learn knowledge regarding the performance of upper-level variables in environmental selection during the evolutionary process.
The model is trained with paired samples received online, which effectively extracts the features of solutions and the relational patterns between solutions, and alleviates the shortage of training data.

$\bullet$ A reference-based ranking mechanism is designed to conveniently evaluate multiple samples instead of pairwise comparing and re-ranking.
In addition, a resampling strategy is employed to improve the quality of generated solutions, which further exploits the contrastive ranking model's capacity in assessing the relative superiority between solutions.

The remainder of the paper is organized as follows:
Section \ref{sec:Preliminaries-and-Related-Work} provides a brief introduction of the preliminaries and related work.
Section \ref{sec:Proposed Method} details the proposed method.
Section \ref{sec:Experimental Study} presents the experiments and analysis.
Finally, Section \ref{sec:Conclusion} concludes the paper with a discussion on future research direction.

\section{Preliminaries and Related Work}
\label{sec:Preliminaries-and-Related-Work}
\subsection{Bilevel Optimization}
A bilevel optimization problem can be formulated as: 
\begin{equation}
\begin{gathered}
\underset{x_u \in X_u}{\textbf{Min}} F\left(x_u, x_l^*\right) \\
\textbf { s.t. } \quad x_l^* \in \underset{x_l \in X_l}{\textbf{argmin}}\left\{f\left(x_u, x_l\right): g_i\left(x_u, x_l\right) \leq 0,\right. \\
i=1,2,\ldots, I\} \\
G_j\left(x_u, x_l\right) \leq 0, j=1,2,\ldots, J
\end{gathered}
\end{equation}
where $F$ and $G$ are the upper-level objective and constraints, respectively, while $f$ and $g$ are the lower-level objective and constraints, respectively.
The objectives and constraints at both levels incorporate the decision variables from both the upper and lower levels.

In addition to the explicit constraints defined by $G$ and $g$, $x_l^*$  indicates the implicit constraints of BLOPs, as only the optimal solution of the lower-level problem with respect to $x_u$ can be considered a valid solution for upper-level evaluation.

\subsection{Related Work}
Bilevel optimization problems have been widely studied due to their complexity and widespread existence, leading to the development of various methods in both mathematical programming and evolutionary algorithms \citep{liu2021investigating}.

Mathematical programming methods usually involve assumptions about the properties of the problems, such as smoothness, convexity, continuity, and differentiability.
A typical strategy is to transform a bilevel problem into a single-level constrained optimization problem using Karush-Kuhn-Tucker (KKT) conditions with Lagrangian and complementarity constraints. 
The simplified problem is then addressed using penalty functions \citep{roghanian2008integrating}, vertex enumeration \citep{tuy1993global}, branch and bound \citep{bard1990branch}, and some mixed-integer solvers \citep{edmunds1991algorithms}.
For instance, Lv \textit{et al.} \citep{lv2007penalty} transformed a linear bilevel programming problem into a single-level problem with an additional penalty by applying the KKT conditions, and then solved it using linear programming.
Trust-region \citep{colson2005trust} and descend methods \citep{vicente1994descent} are also used to solve convex quadratic and nonlinear bilevel programming problems.
The effectiveness of mathematical programming methods is highly dependent on the consistency between the assumptions and the problem’s properties, so these methods are only applicable to relatively simple problems with specific properties mentioned above.

Evolutionary algorithms, recognized for their global search capabilities and versatility, are regarded as promising approaches for dealing with BLOPs and can be classified into three main categories: 1) single-level reduction-based BLEAs, 2) nested structure-based BLEAs and 3) approximation-based BLEAs.

1) Single-level reduction-based BLEAs: This class of methods replaces some convex and regular lower-level problems with optimality conditions, and uses EAs to solve the transformed single-level problems \citep{li2015genetic}.
For instance, after applying KKT-based single-level reduction, Jiang \textit{et al.} \citep{jiang2013application} employed the particle swarm optimization (PSO) algorithm to solve the smoothed nonlinear programming problem transformed by the Chen-Harker-Kanzow-Smale approach. 
Discrete differential evolution \citep{li2016interactive} and estimation of distribution algorithm \citep{wan2014estimation} are also used to solve simplified single-level problems.
Similar to mathematical programming, single-level reduction methods are limited by the assumptions on the problem properties and therefore not suitable for  handling complex problems.

2) Nested structure-based BLEAs: This class of methods follows the hierarchical nature of BLOPs by providing the optimal solution to each lower-level problem derived from the upper-level decision.
Typical EAs, such as genetic algorithm (GA) \citep{wang2011new}, differential evolution (DE) \citep{islam2017enhanced}, PSO \citep{zhao2019nested} and ant colony algorithm \citep{calvete2011bilevel}, have been applied to solve either the upper-level problems or both levels of the problems.
For instance, Huang \textit{et al.} \citep{huang2019jointly} implemented a joint optimization of microgrid configuration and energy consumption scheduling using a bilevel genetic algorithm.
Angelo \textit{et al.} \citep{angelo2015study} applied ant colony optimization at the upper level and differential evolution at the lower level to solve a bilevel production-distribution planning problem.
Nested BLEAs are independent of problem assumptions, thus offering versatility and the ability to handle complex problems including non-differentiable and black-box problems.
However, these methods consume extensive computational resources due to the excessive iterations needed especially in the lower-level optimization process.

3) Approximation-based BLEAs: This class of methods 
constructs approximate models to enhance the search efficiency of nested structure-based BLEAs \citep{sinha2021solving,lin2023classification,singh2019nested}.
Sinha \textit{et al.} \citep{sinha2017evolutionary} proposed a set of methods to approximate the decision relationship between the upper and lower levels, including the reaction set mapping ($\psi$-mapping) and the optimal value function mapping ($\varphi$-mapping).
$\psi$-mapping approximates the optimal solution of the lower-level problem for the upper-level solution, and transforms it into a function of the upper-level solution \citep{sinha2014improved}; $\varphi$-mapping approximates the optimal lower-level function value for the upper-level solution, and transforms it into a constraint of the upper-level problem \citep{sinha2016solving}.
The  pros and cons of these two mappings and their joint mechanism are analyzed in \citep{sinha2020bilevel}.
Islam \textit{et al.} \citep{islam2017surrogate} applied multiple surrogate models, including first- and second-order response surface models and kriging, to approximate the objective and constraint functions at the lower
level.
Kieffer \textit{et al.} \citep{kieffer2019tackling} utilized a genetic programming hyper-heuristic to train a scoring function for solution ranking in the lower-level optimization.
Mamun \textit{et al.} \citep{mamun2021multifidelity} introduced a local augmented Kriging model to predict the fidelity allocated to different lower-level tasks, thereby controlling the number of lower-level iterations.
Approximation-based methods perform well in some specific problems.
However, the selection of approximate models is highly dependent on the prior knowledge of the problems, limiting their generalizability across varying cases. 
Moreover, the uncertainty of the approximate model accuracy may also affect the optimization result.

In recent years, some knowledge sharing-based methods are drawing increasing attention \citep{feng2023multi,jiang2017transfer,jiang2020knee}.
These methods can be categorized under the class of nested structure-based BLEAs, which target to reduce the resource consumption of BLEAs by sharing knowledge especially at the lower level.
For instance, co-evolution-based algorithms select high-quality solutions for information sharing among sub-populations at the lower level or both levels \citep{chaabani2023solving,chaabani2019transfer,chaabani2020co}.
He \textit{et al.} \citep{he2018evolutionary} proposed a bilevel covariance matrix adaptation evolution strategy (BL-CMA-ES), and designed a distribution sharing mechanism to extract a priori knowledge of the lower-level problem from the upper-level optimizer.
Chen \textit{et al.} \citep{chen2021transfer} proposed a transfer learning-based BLEA which processes a set of lower-level problems in a parallel manner, and developed an explicit knowledge transfer strategy.
In \citep{chen2023evolutionary}, multiple lower-level problems for the upper-level population are converted into the task of locating a set of Pareto optimal solutions of a constructed multi-objective optimization problem.
The algorithm leverages the inherent parallelism and implicit similarities in evolutionary multi-objective optimization to improve collaboration and efficiency in solving lower-level problems.
In \citep{huang2023bilevel} and \citep{gupta2015evolutionary}, the lower-level problems corresponding to multiple upper-level
individuals are taken as similar tasks, and the evolutionary multitasking algorithms \citep{gupta2015multifactorial}, \citep{bali2020cognizant} are incorporated
to share valuable information among the lower-level tasks being processed simultaneously.

The aforementioned studies indicate that nested structure-based evolutionary algorithms demonstrate superior problem-solving capability for various BLOPs.
However, their application is limited due to substantial resource consumption.
Although the computational efficiency of these algorithms has been improved through various knowledge transfer strategies, futile efforts on massive unpromising lower-level tasks still result in significant resource waste, which diminishes the gains from these improvements.
To further reduce resource consumption in bilevel optimization algorithms, this work proposes a resource allocation framework for BLEAs which incorporates a contrastive network to identify promising lower-level tasks.

\section{Proposed Method}
\label{sec:Proposed Method}
In this section, we first present the proposed contrastive ranking-based BLEA framework, termed CR-BLEA.
Next, we introduce the contrastive ranking network used for resource allocation, including the structure and motivation of the model.
Finally, we elaborate on the implementation of the strategies including reference-based ranking and resampling.

\begin{figure*}[htbp]
\centering
\includegraphics[width=\textwidth]{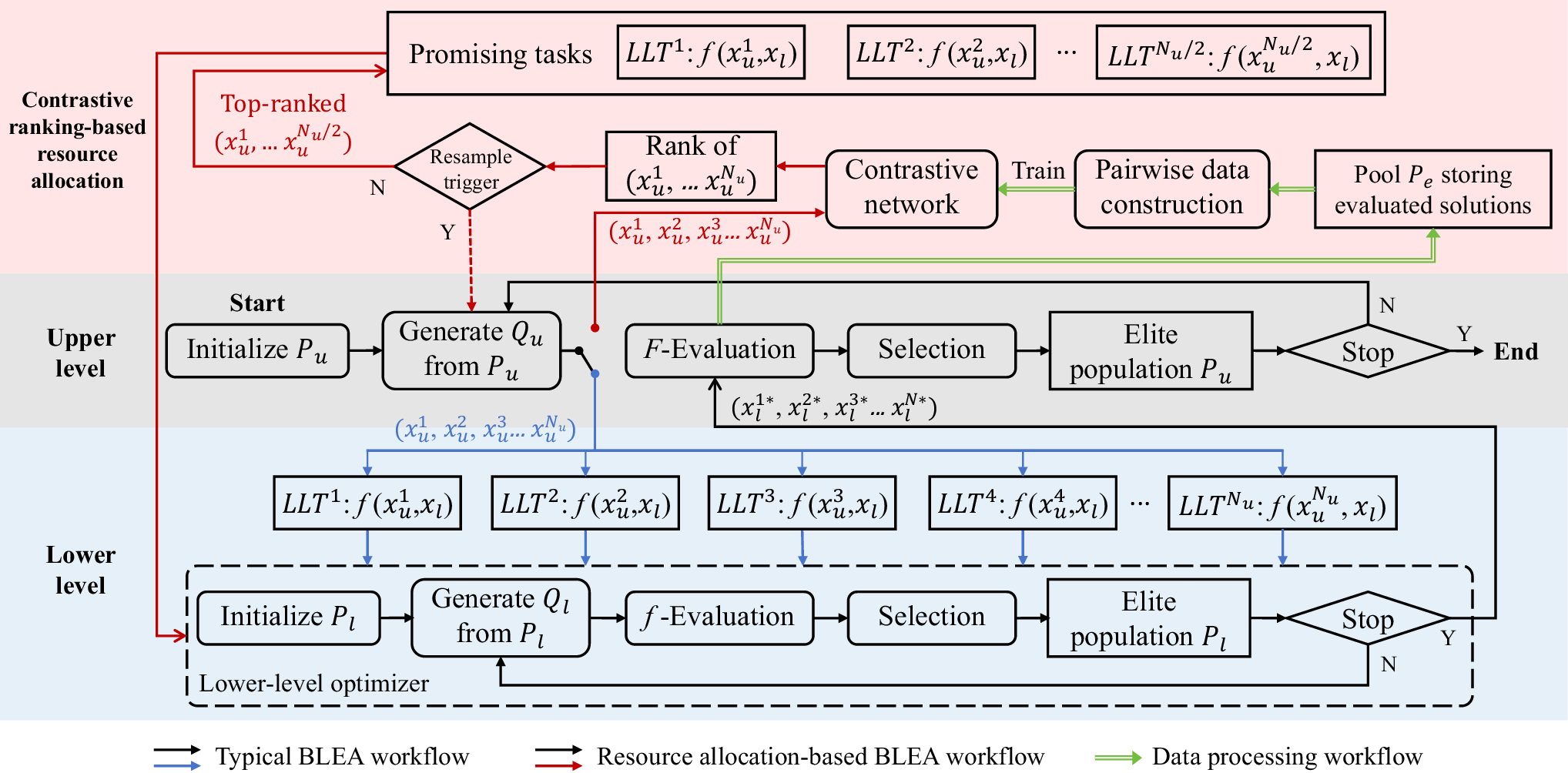}
\caption{Flowchart of the proposed CR-BLEA. The middle section with a gray background represents the upper-level optimization of BLEAs, while the lower section with a blue background represents the lower-level optimization of BLEAs.
The combination of the gray and blue sections depicts a typical process of BLEAs.
The top section with a red background illustrates the proposed contrastive ranking-based resource allocation.
The three sections together describe the overall framework proposed in this paper.
Note that, the absence of a connection from the $Q_u$-generation module to the evaluation module in the upper level reveals that the evaluation of the upper-level variables $x_u$ is indirect.
It requires lower-level optimization to obtain the corresponding $x_l^*$ before upper-level evaluation can be performed.
The diverter switch following $Q_u$-generation module indicate that, at different stages, the algorithm handles multiple new lower-level tasks ($LLTs$) derived from $Q_u$ either in the typical BLEAs manner (blue path) or through contrastive ranking-based resource allocation (red path).}
\label{fig:Flowchart}
\end{figure*}

\begin{algorithm}[tbp]
\caption{Framework of CR-BLEA}
\label{alg:Framework of CR-BLEA}
\KwIn{the bilevel optimization problem.}
\KwOut{the best result $(x_u^*, x_l^*)$.}
\textbf{\emph{// Initialization and the first upper-level iteration}}\;
Initialize upper-level population $P_u \leftarrow \{x_u^1, \dots, x_u^{N_u}\}$\;
Perform lower-level search for all $x_u^i \in P_u$ to obtain $x_l^{i*}$\;
Evaluate $F(x_u^i, x_l^{i*})$ for all $i \in \{1, \dots, N_u\}$  with upper-level FEs\;
        $P_e \gets P_u$\;
    \textit{Flag} $\gets$ 0 \;
\While{the termination condition is not met}{
    \eIf{$|P_e| < N_p $ \textbf{and} Flag == 0}{
        \textbf{\emph{// Typical BLEA process}}\;
        Generate new population $Q_u$ from $P_u$\;
        Perform lower-level search for all $x_u^i \in Q_u$ to obtain $x_l^{i*}$\;
        Evaluate $F(x_u^i, x_l^{i*})$ for all $x_u^i \in Q_u$ \;
        $P_e \gets P_e \cup Q_u$\;
        Update $P_u$ with elite individuals\;
    }{
        \textbf{\emph{// Resource allocation-based process}}\;
        \If{$|P_e| \geq N_p$}{
        $\mathcal{D}\gets \texttt{PDP} (P_e)$ \textbf{\emph{// Algorithm 2}}\;
        Train contrastive network $\mathcal{N}_{cr} \gets \texttt{Train}(\mathcal{N}_{cr},\mathcal{D})$\;
        $P_e \gets \emptyset$\;
}
        $Q_u \gets$ \texttt{PGR} ($\mathcal{N}_{cr},P_u$) \textbf{\emph{// Algorithm 3}}\;
        Perform lower-level search for all $N_u / 2$ individuals in $Q_u$\;
        Evaluate $F(x_u^i, x_l^{i*})$ for all $x_u^i \in Q_u$ \;
        $P_e \gets P_e \cup Q_u$\;
        \textit{Flag} $\gets$ 1 \;
    }
}
\Return $(x_u^*, x_l^*)$ as the best solution found.
\end{algorithm}

\subsection{Framework}
To mitigate the extensive function evaluation waste in BLEAs, this paper proposes a novel resource allocation-based BLEA framework.
As depicted in the overall flowchart in Fig.~\ref{fig:Flowchart}, the middle section with a gray background and the lower section with a blue background collectively represent a typical BLEA process.
To evaluate a new population $Q_u$, all upper-level individuals $x_u = \{x_u^1, \dots, x_u^{N_u}\}$ are sent to the lower-level optimizer via the path indicated by the blue arrows. All lower-level tasks $LLT = \{LLT^1, \dots, LLT^{N_u}\}$ undergo iterative optimization, and the resulting lower-level optimal solutions $x_l^*$ are sent back to the upper level, where they are paired with the corresponding $x_u$ and proceed to the upper-level iteration as indicated by the black arrows.
Such a typical BLEA process is first executed for several iterations in CR-BLEA to accumulate training data.
The evaluated solutions are collected in a pool $P_e$.

Once the number of solutions in $P_e$ is sufficient for training, paired samples are constructed to train a contrastive network that is continuously updated. 
Based on the ranking capability of the trained model, when a new population $Q_u$ is generated in subsequent iterations, all upper-level individuals will be processed along the red path. 
Following typical selection mechanisms in EAs, where the better half is retained, the most promising $N_u/2$ upper-level individuals within $Q_u$ are identified using the model, and only the corresponding lower-level tasks are allocated resources to be processed by the lower-level optimizer.
In addition, by leveraging the model's ability to evaluate the relative advantage between solutions, we can estimate the quality of new population.
If the population quality is considered inferior, a resampling process will be executed.

Throughout the process, the solutions evaluated at the upper level are continuously collected in the pool $P_e$ via the green arrows shown in Fig.~\ref{fig:Flowchart} and used to generate training samples for the model.
The population first follows the typical BLEA process along the blue and black arrows, and then continues to iterate along the red and black arrows in the resource allocation-based process driven by the contrastive ranking network, until the termination condition is met.
The framework of CR-BLEA is illustrated in Algorithm ~\ref{alg:Framework of CR-BLEA}.
Notably, existing nested BLEAs, along with various strategies designed to facilitate the handling of lower-level tasks, can be integrated into the proposed framework.

\subsection{Contrastive model learning}
In the proposed framework, a learning model is constructed to guide resource allocation.
This section first discusses the rationale for designing the model before elaborating on its construction.

\textit{1) Model Preference}:
During the evolutionary process, data regarding the correspondence between $(x_u,x_l^*)$ and $F(x_u,x_l^*)$ can be obtained through  upper-level evaluation.
As analyzed in the introduction, in bilevel optimization problems, $x_l^*$ is a function of $x_u$. 
Therefore, the data obtained is equivalent to the correspondence between $(x_u,\psi(x_u))$ and $F(x_u,\psi(x_u))$, which is further equivalent to the correspondence between $x_u$ and $F(x_u)$.

The purpose of embedding the model in the proposed framework is to learn the potential pattern regarding the correspondence between $x_u$ and $F(x_u)$.
Thus, by utilizing only the upper-level variable values, the model can identify promising upper-level individuals, thereby guiding the allocation of resources.
These resources manifest as opportunities for executing lower-level tasks.
It should be noted that the data used to organize training samples is collected only from solutions evaluated by $F(x_u,x_l^*)$ during the evolutionary process.
Consequently, the model is expected to reliably estimate solution quality under limited data and identify promising solutions from a set of candidates.

Training a classification or regression model is an intuitive approach.
For a classification model, the sample organization involves categorizing the obtained solutions into positive and negative samples.
However, the separation of superior and inferior solutions varies with the evolutionary process.
The population distribution constantly changes, causing the classification boundaries to shift continuously.
In this scenario, a successful iteration may generate new solutions that are superior to the previous ones, but they may all be similar to the superior solutions in the training set, and therefore cannot be correctly separated into two distinct categories.

Regression models are commonly employed in evolutionary computation to predict the function values of solutions, and these predicted values are used to compare the solutions.
However, training a fine-grained regression model with a small dataset is challenging.
The limited amount of data collected during the evolutionary process cannot support accurate function value approximation, and model errors or data noise may lead to inaccurate comparison of solutions.

In essence, what is needed in the proposed framework is not the absolute categorization of solutions as good or bad, nor the absolute function values.
Regardless of the model used, the ultimate goal is to compare the relative advantage of solutions based on the model's predicted results, thereby identifying the better ones in a set of solutions.

Contrastive learning is a machine learning paradigm in which samples are compared against each other to learn feature representations reflecting their similarities or differences, thereby enhancing the model's ability to distinguish between different samples \citep{chen2020simple}.
Inspired by the idea of comparing samples in contrastive learning, we propose to construct a contrastive network to learn the relative relationships between solutions in the training phase and utilize the relational knowledge in the inference phase to estimate the ranking order of a set of solutions.
Note that, the application of the proposed contrastive model in the context of BLEAs differs from the self-supervised nature of existing contrastive learning methods.
Our model not only learns representations but also further assesses their relative advantage.
Therefore, the structure and loss function differ from those of existing contrastive learning methods.

\begin{algorithm}[tbp]
\caption{Paired data preparation (PDP)}
\label{alg:Data preparation}
\KwIn{evaluated solution pool $P_e = \{x_u^1, \dots, x_u^{N}\}$ with corresponding fitness values $\{F(x_u^1,x_l^{1*}), \dots, F(x_u^N,x_l^{N*})\}$.}
\KwOut{training set $\mathcal{D}$: $\{\langle[x_u^i,x_u^j],l_{i,j}\rangle|i,j = 1, \dots, N, i \neq j\}$.}
\For{$i = 1$ to $N$}
{\For{$j = 1$ to $i-1$}
{
$l_{i,j} = sgn(F(x_u^j,x_l^{j*})-F(x_u^i,x_l^{i*}))$\;
$\mathcal{D} = \mathcal{D} \cup  {\langle[x_u^i,x_u^j],\frac{l + 1}{2}\rangle}$ \;
$\mathcal{D} = \mathcal{D} \cup  {\langle[x_u^j,x_u^i],\frac{-l + 1}{2}\rangle}$ \;
}
}
\Return training set $\mathcal{D}$\;
\end{algorithm}

Compared with classification and regression models, the proposed model has the following advantages:
1) The contrastive model aligns with the contrastive requirement of the framework, specifically to recognize the comparative relationships between solutions.
Besides, the designed structure allows for ranking a set of solutions without pairwise comparisons, making it more straightforward.
2) The pattern of relative relationships between solutions is easier to learn than the absolute function value mappings.
Moreover, the pattern generalizes well across solutions at different evolutionary stages, which is not interfered by changes in population distribution.
3) With $N$ original samples regarding $x_u$ and $F(x_u)$, $N(N-1)$ training samples can be generated based on pairwise relationships. Such data augmentation can facilitate model training.

\textit{2) Model construction}:
The dataset $\mathcal{D}$ used for model training is organized by pairing samples from the pool $P_e$ used to accumulate evaluated solutions:
\begin{equation}
\begin{aligned}
\mathcal{D} & =\left\{\left(\left(x_u^i, x_u^j\right), l_{i,j}\right) \mid x_u^i, x_u^j \in P_e\right\} \\
l_{i,j} & = \begin{cases}1, & F(x_u^i,x_l^{i*})<F(x_u^j,x_l^{j*}) \\
0, & F(x_u^i,x_l^{i*})>F(x_u^j,x_l^{j*})\end{cases}
\end{aligned}
\end{equation}
where $l_{i,j}$ = 1 indicates that solution $x_u^i$ outperforms $x_u^j$ with a better (i,e., smaller) fitness $F(x_u,x_l^*)$, thereby being considered a more promising solution.
The pseudo-code for the sample generation is provided in Algorithm \ref{alg:Data preparation}.

\begin{figure}[htbp]
  \centering
  \subfigure[Contrastive ranking network]
  {
    \label{fig:ranknet}
    \includegraphics[width=0.38733\linewidth]{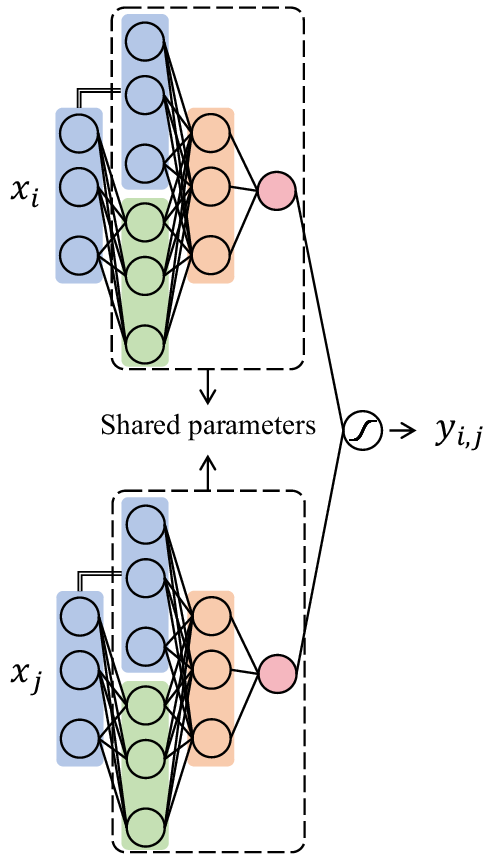}
  }
  \subfigure[Subnet]
  {
    \label{fig:subnet}
    \includegraphics[width=0.49467\linewidth]{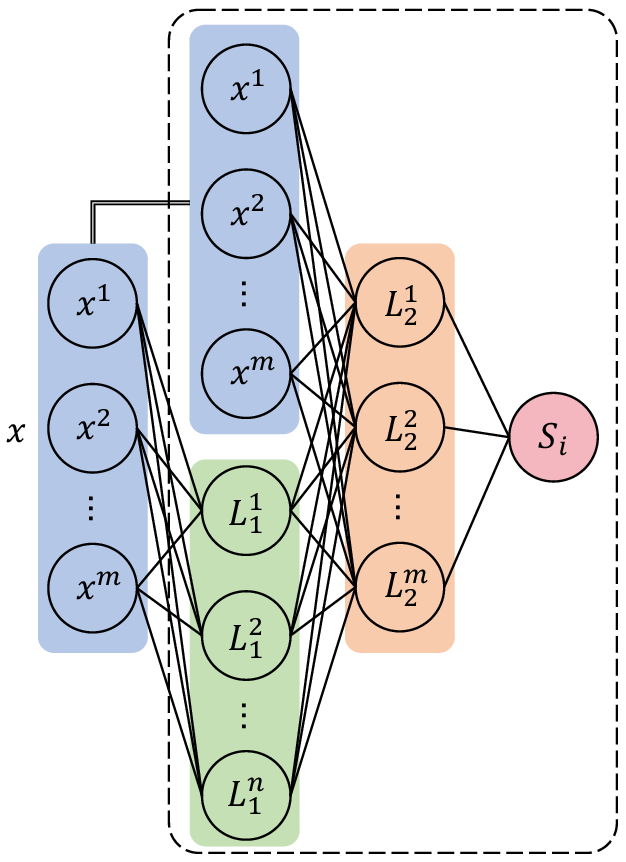}
  }
  \caption{Structure of the proposed contrastive ranking network.} 
  \label{net structure}
\end{figure}

The structure of the proposed contrastive ranking network named $\mathcal{N}_{cr}$ is illustrated in Fig. \ref{fig:ranknet}.
The model comprises two sub-networks with shared parameters.
To learn the relationship between the solution pairs $x_u^i$ and $x_u^j$, the variable information of the two individuals are sent into the sub-networks respectively. The sub-networks extract features and provide the representations of the two solutions, denoted as $S_i$ and $S_j$, respectively.
Subsequently, the difference between $S_i$ and $S_j$ is
calculated with a sigmoid function as:
\begin{equation}
\hat{y}_{i,j}=\sigma(S_i - S_j)
\end{equation}

The output $\hat{y}_{i,j}$ of the network ranges between [0,1], indicating the probability that individual $x_u^i$ is superior to $x_u^j$.
Analogous to a contrastive siamese network \citep{khorram2022contrastive}, the sub-networks with shared parameters provide a fair comparison by extracting the features of both solutions in the same way.
In addition, the number of parameters the entire model needs to learn is reduced, which further relieves the risk of over-fitting together with paired data augmentation.

Fig. \ref{fig:subnet} depicts the structure of the sub-network.
The sub-network takes the $m$-dimensional variables of an upper-level individual $x_u^i$ as input, and then extracts features with several fully connected layers, where ReLU is used as the activation function.
As analyzed in the model preference, the feature extraction in the contrastive network involves the relation between $x_u$ and $F(x_u)$, which is inferred from $F(x_u,x_l^*)$ and $F(x_u,\psi(x_u))$.
To simulate this potential relation, a quasi-residual structure is designed in the sub-network.
The $m$-dimensional variable feature of $x_u$ is first connected to $n$ neurons ($n$ is the dimension of the lower-level decision variables) through a fully connected layer to simulate the mapping $\psi(x_u)$ from $x_u$ to $x_l^*$.
The $n$-dimensional output will enter the subsequent fully connected layer together with the $m$-dimensional original feature, and finally obtain the output $S_i$ of the sub-network.
In the subsequent experiments, the structure and the number of neurons in the sub-network are consistent with those described in Fig. \ref{fig:subnet}, where $m$ and $n$ denote the dimensions of the upper- and lower-level decision variables, respectively.

To learn the parameters of the contrastive network, binary cross-entropy (BCE) loss is used to quantify the difference between $\hat{y}_{i,j}$ and $l_{i,j}$.
The Adam optimizer with a learning rate of 0.1 is used to minimize the loss via back-propagation.
The frequency of network updates is determined by the number of evaluated solutions in the pool $P_e$.
Once the accumulated solutions reach a number $N_p$ that is sufficient to be paired and organize a dataset with $N_p(N_p-1)$ samples as ten times the size of the network's trainable parameters, a new training dataset $\mathcal{D}$ is constructed and used to update the contrastive network.

\begin{algorithm}[tbp]
\caption{Population generation with ranking and resampling (PGR)}
\label{alg:Generate_and_Select}
\KwIn{contrastive network $\mathcal{N}_{cr}$, population $P_u$ with corresponding ranking score.}
\KwOut{population $Q_u$ with $N_u/2$ top-ranked solutions.}

Generate new population $Q_u$ from $P_u$\;
\ForEach{$x_u^i \in Q_u$}{
    Calculate ranking score $r(x_u^{i})$ using $\mathcal{N}_{cr}^r$ derived from $\mathcal{N}_{cr}$\;
}
$Q_u \gets$ $N_u/2$ top-ranked individuals in $Q_u$ based on ranking scores\;

\If{$\max(r_i \mid x_u^i \in Q_u) < \max(r_j \mid x_u^j \in P_u)$}{
    Generate new population $R_u$ from $P_u$\;
    \ForEach{$x_u^i \in R_u$}{
        Calculate ranking score $r(x_u^{i})$ using $\mathcal{N}_{cr}^R$\;
    }
    $Q_u \gets$ $N_u/2$ top-ranked individuals in $Q_u \cup R_u$ based on ranking scores\;
}
\Return $Q_u$ with $N_u/2$ top-ranked individuals.
\end{algorithm}

\subsection{Contrastive model Usage}
\textit{1) Reference-based Ranking}:
Based on the ability of $\mathcal{N}_{cr}$ to infer the relative relation between solutions, two newly generated solutions can be fed into the network for comparison to identify the more promising one.
However, when dealing with a set of solutions, obtaining the ranking of all solutions through pairwise comparison comes with limitations.
On one hand, the $\frac{N(N-1)}{2}$ results of pairwise comparisons among all solutions need to be reordered.
On the other hand, transitivity conflict may occur, where the predicted results could lead to a contradictory cycle as $x_u^a$ being better than $x_u^b$, $x_u^b$ better than $x_u^c$, but $x_u^a$ not better than $x_u^c$ \citep{huang2022contrastive}.

To address these limitations, a reference-based ranking strategy is employed for handling a set of solutions.
In the inference phase, a variant model called $\mathcal{N}_{cr}^R$ is constructed, which retains only half of the model related to the first individual $x_u^i$, while fixing the output $S_j$ of the other half at 0.
The intention of $\mathcal{N}_{cr}^R$ is to compare the input variable $x_u^i$ with a reference variable $x_u^r$ that results in $S_r$ = 0.
All solutions $x_u^k$ in $Q_u$ are inferred using $\mathcal{N}_{cr}^R$, and the results indicate the probability that $x_u^k$ is more promising than $x_u^r$.
This process is equivalent to measuring the advantage of different solutions over the same reference solution.
Based on the advantage value recorded as the ranking score $r(x_u^i)$, the ranking of all solutions in $Q_u$ can be obtained.
With the reference-based ranking strategy, the model can learn pairwise contrastive knowledge in the training phase while avoiding the complexity of pairwise re-ranking in the inference phase.

\textit{2) Resampling Strategy}: The reference-based ranking can identify the more promising parts of a set of newly generated solutions based on the relative relation, thus providing guidance for resource allocation in BLEAs.
However, it is incapable of estimating the absolute quality of solutions.
This leads to the limitation that if the offspring solutions offer no actual improvement over the parent population, the lower-level optimization resources invested in them are still wasted, even if the relatively more promising ones have been selected.

To ensure the improvement of offspring generation, the potential of the contrastive ranking model is further leveraged by implementing a resampling strategy.
Based on the same reference $S_r$ = 0, the ranking score of the best solution in the offspring population is compared with that of the best solution in the parent population.
If the best ranking score of the offspring is lower than that of the parent population, it implies that no better solution has been generated in the offspring.
In this case, resampling is triggered, as the offspring population is regenerated to maintain improvement in the quality of upper-level individuals in the evolutionary process, thereby further enhancing the effectiveness of resource allocation.
If the evolution of the upper population stagnates, resampling can be performed multiple times.
In the proposed framework, we only conduct a trial implementation that for each population, resampling is performed once at most.
Promising individuals are then selected from all solutions of the two samplings based on the predicted ranking scores.
The pseudo-code for the population generation with ranking and resampling is provided in Algorithm \ref{alg:Generate_and_Select}.

\subsection{Computational Complexity Analysis}
The complexity of CR-BLEA framework varies with the employed bilevel optimizer; therefore, we consider CR‐TLEA‐CMA‐ES as an example for brevity.
As indicated in \citep{chen2021transfer}, the computational complexity of TLEA‐CMA‐ES is $O(N \cdot d^3)$, where $N$ and $d$ denote the upper level population size and the number of lower‐level variables, respectively.
In particular, the number of tasks that need to be handled in CR-TLEA‐CMA‐ES can be reduced to $N/2$ through resource allocation.
Regarding the resource allocation mechanism, the computational cost is mainly attributed to data organization and network training.
One network update entails $O(N^2)$ operations for organizing the samples, while training the contrastive network requires a complexity of $O(N \cdot d \cdot q)$, where $q$ denotes the number of neurons in the hidden layer. 
As $d \cdot q$ significantly exceeds the population size $N$, the total operations of network update are $O(N \cdot d \cdot q + N^2)$ = $O(N \cdot d \cdot q)$.

\section{Experimental Study}
\label{sec:Experimental Study}
\subsection{Benchmark Problems and Performance Indicators}
All algorithms considered are examined on the widely recognized SMD test suite \citep{sinha2014test} with 12 bilevel problems and TP test suite \citep{sinha2017evolutionary} with 10 bilevel problems.
SMD test suite includes challenges in terms of controlled convergence, interaction between upper and lower levels, multimodality and the constraints encountered in the problems, while most of the functions in TP problems are linear or quadratic with constraints included.
All problems in SMD are tested with the upper- and lower- variable dimensions set to ($m$ = 2, $n$ = 3), while the variable dimensions of each problem in TP are predefined.
The real optimal function values $(F^r,f^r)$ of the problems are known and used for performance evaluation.

In the numerical experiments, accuracy and the number of function evaluations are used as indicators to assess the algorithm's problem-solving ability and resource consumption:

$\bullet$ The accuracy measures the deviation between the real optimal function values $(F^r,f^r)$ and the best results $(F^*,f^*)$ found when the algorithm terminates.
The accuracy in BLEAs is calculated at both levels as $Acc_u$ = $\left|F^r-F^*\right|$ and $Acc_l$ = $\left|f^r-f^*\right|$.
For ease of comparison, the results with accuracy below $1e^{-6}$ are marked as $1e^{-6}$ to filter out subtle differences, as is commonly treated in BLEA studies \citep{he2018evolutionary}, 
\citep{chen2023evolutionary}, \citep{islam2017surrogate}, \citep{huang2020framework}.

$\bullet$ The number of function evaluations consumed is recorded at both the upper and lower levels and denoted as $FEs_u$ and $FEs_l$, respectively.
To assess the overall resource consumption, the total number of function evaluations is also considered as $FEs_t = FEs_u + FEs_l$.
The resource-saving rate of algorithm A against algorithm B is calculated as:
\begin{equation}
R_{rs}^{A,B} = \frac{FEs_t^B - FEs_t^A}{FEs_t^B} \times 100\%
\end{equation}

\subsection{Compared Algorithms and Parameter Settings}
The following five advanced BLEAs are considered for performance comparison against the variants incorporated with the proposed design framework:

$\bullet$ TLEA-CMA-ES \citep{chen2021transfer}: a transfer learning-based BLEA that handles multiple lower-level tasks in parallel and transfers the distribution knowledge from source tasks to target tasks.

$\bullet$ TLEA-DE \citep{chen2021transfer}: a BLEA that adopts the same parallel transfer strategy as TLEA-CMA-ES, with DE as the single-level optimizer.

$\bullet$ BL-CMA-ES \citep{he2018evolutionary}: a covariance matrix adaptation-based BLEA that employs CMA-ES at both levels, and shares the search distribution from the upper level to the lower level as prior knowledge.

$\bullet$ MOTEA \citep{chen2023evolutionary}: a multi-objective transformation-based BLEA which converts multiple lower-level optimization processes into a single multi-objective optimization task.

$\bullet$ BOC \citep{huang2023bilevel}: a collaboration-based BLEA that uses a population to solve the lower-level optimization tasks for all upper-level individuals in a single run with an information-sharing mechanism.

To verify the effectiveness of proposed framework,
these algorithms are integrated into CR-BLEA and their names are modified by adding "CR-" at the front.
For instance, the variant of BOC is denoted as CR-BOC.

The parameters of the tested algorithms and their variants remain unchanged from their default settings.
For all algorithms, the upper-level population size and the lower-level population size are set to $4+\lfloor\ln(m+n)\rfloor$ and $4+\lfloor\ln(n)\rfloor$, respectively.
The upper-level optimization terminates if 1) the number of upper-level FEs consumed exceeds $FEs_u^{max}$ or 2) the elitist upper-level objective value varies within a range of less than $1e^{-6}$ over the last $FEs_u^{var}$ or 3) the accuracy of the elitist upper-level objective value is less than $1e^{-6}$ from the known optimal solution.
The lower-level optimization terminates if 1) the number of lower-level FEs consumed exceeds $FEs_l^{max}$ or 2) the elitist lower-level objective value varies within a range of less than 1$e-5$ over the last $FEs_l^{var}$. 
The termination conditions for SMD and TP problems are set the same as $FEs_u^{max}$ = 2500, $FEs_u^{var}$ = 350, $FEs_l^{max}$ = 250 and $FEs_l^{var}$ = 25.
For BOC and CR-BOC, the lower-level optimization stops if the number of generations reaches 18, as is consistent with the original paper \citep{huang2023bilevel}.
All algorithms are run 21 times on each test instance independently.

\subsection{Comparison Study}
The comparison results between the five tested algorithms and their variants regarding the accuracy and the number of function evaluations on SMD problems are presented in Tables \ref{tab:TLEACMAES_vs_CRTLEACMAES} - \ref{tab:BOC_vs_CRBOC}.
For an intuitive comparison, the better results in each case are highlighted in a gray background, and the Wilcoxon test is performed at a 0.05 significance level to indicate the difference between comparison results.

\begin{table*}[htbp]
  \centering
  \caption{Performance comparison between TLEA-CMA-ES and CR-TLEA-CMA-ES regarding the median results of the accuracy and the number of FEs on SMD problems in 21 runs.}
  \label{tab:TLEACMAES_vs_CRTLEACMAES}
\resizebox{\textwidth}{!}{
    \begin{tabular}{ccccccccccrc}
    \toprule
    \multirow{2}[4]{*}{Probelm} & \multicolumn{5}{c}{TLEA-CMA-ES}       &       & \multicolumn{5}{c}{CR-TLEA-CMA-ES} \\
\cmidrule{2-6}\cmidrule{8-12}          & $Acc_u$  & $Acc_l$  & $FEs_u$  & $FEs_l$  & $FEs_t$  &       & $Acc_u$  & $Acc_l$  & \multicolumn{1}{c}{$FEs_u$} & \multicolumn{1}{c}{$FEs_l$} & $FEs_t$ ($R_{rs}$) \\
\cmidrule{1-6}\cmidrule{8-12}    SMD1  & \multicolumn{1}{{c}}{\cellcolor[rgb]{.851,.851,.851}1.00E-06 ($\approx$)} & \multicolumn{1}{{c}}{\cellcolor[rgb]{.851,.851,.851}1.00E-06 ($\approx$)} & \multicolumn{1}{{c}}{3.21E+02 (+)} & \multicolumn{1}{{c}}{2.00E+04 (+)} & \multicolumn{1}{{c}}{2.03E+04 (+)} &       & \cellcolor[rgb]{.851,.851,.851}1.00E-06 & \cellcolor[rgb]{.851,.851,.851}1.00E-06 & \multicolumn{1}{c}{\cellcolor[rgb]{.851,.851,.851}1.77E+02} & \multicolumn{1}{c}{\cellcolor[rgb]{.851,.851,.851}1.26E+04} & \cellcolor[rgb]{.851,.851,.851}1.28E+04 (37.1\%) \\
\cmidrule{1-6}\cmidrule{8-12}    SMD2  & \multicolumn{1}{{c}}{\cellcolor[rgb]{.851,.851,.851}1.00E-06 ($\approx$)} & \multicolumn{1}{{c}}{\cellcolor[rgb]{.851,.851,.851}1.08E-06 ($\approx$)} & \multicolumn{1}{{c}}{2.84E+02 (+)} & \multicolumn{1}{{c}}{1.98E+04 (+)} & \multicolumn{1}{{c}}{2.01E+04 (+)} &       & \cellcolor[rgb]{.851,.851,.851}1.00E-06 & 1.16E-06 & \multicolumn{1}{c}{\cellcolor[rgb]{.851,.851,.851}1.72E+02} & \multicolumn{1}{c}{\cellcolor[rgb]{.851,.851,.851}1.24E+04} & \cellcolor[rgb]{.851,.851,.851}1.25E+04 (37.7\%) \\
\cmidrule{1-6}\cmidrule{8-12}    SMD3  & \multicolumn{1}{{c}}{\cellcolor[rgb]{.851,.851,.851}1.00E-06 ($\approx$)} & \multicolumn{1}{{c}}{\cellcolor[rgb]{.851,.851,.851}1.00E-06 ($\approx$)} & \multicolumn{1}{{c}}{2.99E+02 (+)} & \multicolumn{1}{{c}}{2.04E+04 (+)} & \multicolumn{1}{{c}}{2.07E+04 (+)} &       & \cellcolor[rgb]{.851,.851,.851}1.00E-06 & \cellcolor[rgb]{.851,.851,.851}1.00E-06 & \multicolumn{1}{c}{\cellcolor[rgb]{.851,.851,.851}1.85E+02} & \multicolumn{1}{c}{\cellcolor[rgb]{.851,.851,.851}1.35E+04} & \cellcolor[rgb]{.851,.851,.851}1.37E+04 (34.0\%) \\
\cmidrule{1-6}\cmidrule{8-12}    SMD4  & \multicolumn{1}{{c}}{\cellcolor[rgb]{.851,.851,.851}1.00E-06 ($\approx$)} & \multicolumn{1}{{c}}{1.98E-06 ($\approx$)} & \multicolumn{1}{{c}}{3.47E+02 (+)} & \multicolumn{1}{{c}}{2.24E+04 (+)} & \multicolumn{1}{{c}}{2.28E+04 (+)} &       & \cellcolor[rgb]{.851,.851,.851}1.00E-06 & \cellcolor[rgb]{.851,.851,.851}1.89E-06 & \multicolumn{1}{c}{\cellcolor[rgb]{.851,.851,.851}2.12E+02} & \multicolumn{1}{c}{\cellcolor[rgb]{.851,.851,.851}1.35E+04} & \cellcolor[rgb]{.851,.851,.851}1.37E+04 (39.9\%) \\
\cmidrule{1-6}\cmidrule{8-12}    SMD5  & \multicolumn{1}{{c}}{\cellcolor[rgb]{.851,.851,.851}1.00E-06 ($\approx$)} & \multicolumn{1}{{c}}{1.75E-06 ($\approx$)} & \multicolumn{1}{{c}}{3.74E+02 (+)} & \multicolumn{1}{{c}}{2.16E+04 (+)} & \multicolumn{1}{{c}}{2.19E+04 (+)} &       & \cellcolor[rgb]{.851,.851,.851}1.00E-06 & \cellcolor[rgb]{.851,.851,.851}1.17E-06 & \multicolumn{1}{c}{\cellcolor[rgb]{.851,.851,.851}2.09E+02} & \multicolumn{1}{c}{\cellcolor[rgb]{.851,.851,.851}1.33E+04} & \cellcolor[rgb]{.851,.851,.851}1.35E+04 (38.2\%) \\
\cmidrule{1-6}\cmidrule{8-12}    SMD6  & \multicolumn{1}{{c}}{\cellcolor[rgb]{.851,.851,.851}1.00E-06 ($\approx$)} & \multicolumn{1}{{c}}{\cellcolor[rgb]{.851,.851,.851}1.00E-06 ($\approx$)} & \multicolumn{1}{{c}}{4.06E+02 (+)} & \multicolumn{1}{{c}}{2.39E+04 (+)} & \multicolumn{1}{{c}}{2.43E+04 (+)} &       & \cellcolor[rgb]{.851,.851,.851}1.00E-06 & \cellcolor[rgb]{.851,.851,.851}1.00E-06 & \multicolumn{1}{c}{\cellcolor[rgb]{.851,.851,.851}3.86E+02} & \multicolumn{1}{c}{\cellcolor[rgb]{.851,.851,.851}2.22E+04} & \cellcolor[rgb]{.851,.851,.851}2.26E+04 (6.9\%) \\
\cmidrule{1-6}\cmidrule{8-12}    SMD7  & \multicolumn{1}{{c}}{\cellcolor[rgb]{.851,.851,.851}1.00E-06 ($\approx$)} & \multicolumn{1}{{c}}{1.03E-06 ($\approx$)} & \multicolumn{1}{{c}}{3.14E+02 (+)} & \multicolumn{1}{{c}}{2.16E+04 (+)} & \multicolumn{1}{{c}}{2.19E+04 (+)} &       & \cellcolor[rgb]{.851,.851,.851}1.00E-06 & \cellcolor[rgb]{.851,.851,.851}1.00E-06 & \multicolumn{1}{c}{\cellcolor[rgb]{.851,.851,.851}1.76E+02} & \multicolumn{1}{c}{\cellcolor[rgb]{.851,.851,.851}1.25E+04} & \cellcolor[rgb]{.851,.851,.851}1.27E+04 (41.8\%) \\
\cmidrule{1-6}\cmidrule{8-12}    SMD8  & \multicolumn{1}{{c}}{\cellcolor[rgb]{.851,.851,.851}1.00E-06 ($\approx$)} & \multicolumn{1}{{c}}{\cellcolor[rgb]{.851,.851,.851}1.00E-06 ($\approx$)} & \multicolumn{1}{{c}}{1.40E+03 (+)} & \multicolumn{1}{{c}}{6.63E+04 (+)} & \multicolumn{1}{{c}}{6.77E+04 (+)} &       & \cellcolor[rgb]{.851,.851,.851}1.00E-06 & \cellcolor[rgb]{.851,.851,.851}1.00E-06 & \multicolumn{1}{c}{\cellcolor[rgb]{.851,.851,.851}6.85E+02} & \multicolumn{1}{c}{\cellcolor[rgb]{.851,.851,.851}3.55E+04} & \cellcolor[rgb]{.851,.851,.851}3.62E+04 (46.5\%) \\
\cmidrule{1-6}\cmidrule{8-12}    SMD9  & \multicolumn{1}{{c}}{\cellcolor[rgb]{.851,.851,.851}1.00E-06 ($\approx$)} & \multicolumn{1}{{c}}{1.54E-06 ($\approx$)} & \multicolumn{1}{{c}}{3.00E+02 (+)} & \multicolumn{1}{{c}}{1.89E+04 (+)} & \multicolumn{1}{{c}}{1.92E+04 (+)} &       & \cellcolor[rgb]{.851,.851,.851}1.00E-06 & \cellcolor[rgb]{.851,.851,.851}1.00E-06 & \multicolumn{1}{c}{\cellcolor[rgb]{.851,.851,.851}1.90E+02} & \multicolumn{1}{c}{\cellcolor[rgb]{.851,.851,.851}1.27E+04} & \cellcolor[rgb]{.851,.851,.851}1.29E+04 (32.8\%) \\
\cmidrule{1-6}\cmidrule{8-12}    SMD10 & \multicolumn{1}{{c}}{1.60E+01 (+)} & \multicolumn{1}{{c}}{1.60E+01 (+)} & \multicolumn{1}{{c}}{2.50E+03 (+)} & \multicolumn{1}{{c}}{8.91E+04 (+)} & \multicolumn{1}{{c}}{9.16E+04 (+)} &       & \cellcolor[rgb]{.851,.851,.851}8.28E-03 & \cellcolor[rgb]{.851,.851,.851}2.14E-03 & \multicolumn{1}{c}{\cellcolor[rgb]{.851,.851,.851}1.18E+03} & \multicolumn{1}{c}{\cellcolor[rgb]{.851,.851,.851}5.59E+04} & \cellcolor[rgb]{.851,.851,.851}5.71E+04 (37.7\%) \\
\cmidrule{1-6}\cmidrule{8-12}    SMD11 & \multicolumn{1}{{c}}{1.28E-03 (+)} & \multicolumn{1}{{c}}{1.64E-03 (+)} & \multicolumn{1}{{c}}{2.50E+03 (+)} & \multicolumn{1}{{c}}{1.26E+05 (+)} & \multicolumn{1}{{c}}{1.28E+05 (+)} &       & \cellcolor[rgb]{.851,.851,.851}1.00E-06 & \cellcolor[rgb]{.851,.851,.851}8.61E-05 & \multicolumn{1}{c}{\cellcolor[rgb]{.851,.851,.851}1.64E+03} & \multicolumn{1}{c}{\cellcolor[rgb]{.851,.851,.851}7.69E+04} & \cellcolor[rgb]{.851,.851,.851}7.86E+04 (38.7\%) \\
\cmidrule{1-6}\cmidrule{8-12}    SMD12 & \multicolumn{1}{{c}}{\cellcolor[rgb]{.851,.851,.851}1.00E-06 ($\approx$)} & \multicolumn{1}{{c}}{4.14E-06 ($\approx$)} & \multicolumn{1}{{c}}{9.13E+02 (+)} & \multicolumn{1}{{c}}{4.68E+04 (+)} & \multicolumn{1}{{c}}{4.77E+04 (+)} &       & \cellcolor[rgb]{.851,.851,.851}1.00E-06 & \cellcolor[rgb]{.851,.851,.851}3.00E-06 & \multicolumn{1}{c}{\cellcolor[rgb]{.851,.851,.851}8.73E+02} & \multicolumn{1}{c}{\cellcolor[rgb]{.851,.851,.851}4.27E+04} & \cellcolor[rgb]{.851,.851,.851}4.36E+04 (8.6\%) \\
\cmidrule{1-6}\cmidrule{8-12}    +/$\approx$/- & 2/10/0 & 2/10/0 & 12/0/0 & 12/0/0 & 12/0/0 &       &       &       & \multicolumn{2}{r}{Average $R_{rs}$} & 33.3\% \\
    \bottomrule
    \end{tabular}%
}
     \begin{tablenotes}
        \scriptsize
        \item  The better results in each case are highlighted in a gray background. "+", "$\approx$" and "-" indicate that the variant performs significantly better than, equivalent to, and worse than the original algorithm, respectively. The values in parentheses present the relative percentage reduction of $FEs_t$ consumed by the variant. The same symbols are applied to other tables.
      \end{tablenotes}
\end{table*}%

\begin{table*}[htbp]
  \centering
  \caption{Performance comparison between TLEA-DE and CR-TLEA-DE regarding the median results of the accuracy and the number of FEs on SMD problems in 21 runs.}
    \label{tab:TLEADE_vs_CRTLEADE}
    \resizebox{\textwidth}{!}{    \begin{tabular}{ccccccccccrc}
    \toprule
    \multirow{2}[4]{*}{Probelm} & \multicolumn{5}{c}{TLEA-DE}           &       & \multicolumn{5}{c}{CR-TLEA-DE} \\
\cmidrule{2-6}\cmidrule{8-12}          & $Acc_u$  & $Acc_l$  & $FEs_u$  & $FEs_l$  & $FEs_t$  &       & $Acc_u$  & $Acc_l$  & \multicolumn{1}{c}{$FEs_u$} & \multicolumn{1}{c}{$FEs_l$} & $FEs_t$ ($R_{rs}$)\\
\cmidrule{1-6}\cmidrule{8-12}    SMD1  & \multicolumn{1}{{c}}{\cellcolor[rgb]{.851,.851,.851}1.00E-06 ($\approx$)} & \multicolumn{1}{{c}}{\cellcolor[rgb]{.851,.851,.851}1.00E-06 ($\approx$)} & \multicolumn{1}{{c}}{3.51E+02 (+)} & \multicolumn{1}{{c}}{1.30E+04 (+)} & \multicolumn{1}{{c}}{1.33E+04 (+)} &       & \cellcolor[rgb]{.851,.851,.851}1.00E-06 & \cellcolor[rgb]{.851,.851,.851}1.00E-06 & \multicolumn{1}{c}{\cellcolor[rgb]{.851,.851,.851}1.99E+02} & \multicolumn{1}{c}{\cellcolor[rgb]{.851,.851,.851}8.25E+03} & \cellcolor[rgb]{.851,.851,.851}8.45E+03 (36.6\%) \\
\cmidrule{1-6}\cmidrule{8-12}    SMD2  & \multicolumn{1}{{c}}{\cellcolor[rgb]{.851,.851,.851}1.00E-06 ($\approx$)} & \multicolumn{1}{{c}}{3.01E-06 ($\approx$)} & \multicolumn{1}{{c}}{3.02E+02 (+)} & \multicolumn{1}{{c}}{1.14E+04 (+)} & \multicolumn{1}{{c}}{1.17E+04 (+)} &       & \cellcolor[rgb]{.851,.851,.851}1.00E-06 & \cellcolor[rgb]{.851,.851,.851}1.06E-06 & \multicolumn{1}{c}{\cellcolor[rgb]{.851,.851,.851}2.05E+02} & \multicolumn{1}{c}{\cellcolor[rgb]{.851,.851,.851}7.69E+03} & \cellcolor[rgb]{.851,.851,.851}7.90E+03 (32.4\%) \\
\cmidrule{1-6}\cmidrule{8-12}    SMD3  & \multicolumn{1}{{c}}{\cellcolor[rgb]{.851,.851,.851}1.00E-06 ($\approx$)} & \multicolumn{1}{{c}}{1.19E-06 ($\approx$)} & \multicolumn{1}{{c}}{3.50E+02 (+)} & \multicolumn{1}{{c}}{1.32E+04 (+)} & \multicolumn{1}{{c}}{1.35E+04 (+)} &       & \cellcolor[rgb]{.851,.851,.851}1.00E-06 & \cellcolor[rgb]{.851,.851,.851}1.00E-06 & \multicolumn{1}{c}{\cellcolor[rgb]{.851,.851,.851}2.05E+02} & \multicolumn{1}{c}{\cellcolor[rgb]{.851,.851,.851}8.43E+03} & \cellcolor[rgb]{.851,.851,.851}8.63E+03 (36.1\%) \\
\cmidrule{1-6}\cmidrule{8-12}    SMD4  & \multicolumn{1}{{c}}{\cellcolor[rgb]{.851,.851,.851}1.00E-06 ($\approx$)} & \multicolumn{1}{{c}}{\cellcolor[rgb]{.851,.851,.851}1.67E-06 ($\approx$)} & \multicolumn{1}{{c}}{3.75E+02 (+)} & \multicolumn{1}{{c}}{1.38E+04 (+)} & \multicolumn{1}{{c}}{1.42E+04 (+)} &       & \cellcolor[rgb]{.851,.851,.851}1.00E-06 & 3.94E-06 & \multicolumn{1}{c}{\cellcolor[rgb]{.851,.851,.851}2.31E+02} & \multicolumn{1}{c}{\cellcolor[rgb]{.851,.851,.851}8.94E+03} & \cellcolor[rgb]{.851,.851,.851}9.17E+03 (35.3\%) \\
\cmidrule{1-6}\cmidrule{8-12}    SMD5  & \multicolumn{1}{{c}}{\cellcolor[rgb]{.851,.851,.851}1.00E-06 ($\approx$)} & \multicolumn{1}{{c}}{\cellcolor[rgb]{.851,.851,.851}2.12E-06 ($\approx$)} & \multicolumn{1}{{c}}{3.88E+02 (+)} & \multicolumn{1}{{c}}{1.33E+04 (+)} & \multicolumn{1}{{c}}{1.36E+04 (+)} &       & \cellcolor[rgb]{.851,.851,.851}1.00E-06 & 2.13E-06 & \multicolumn{1}{c}{\cellcolor[rgb]{.851,.851,.851}2.25E+02} & \multicolumn{1}{c}{\cellcolor[rgb]{.851,.851,.851}8.46E+03} & \cellcolor[rgb]{.851,.851,.851}8.68E+03 (36.4\%) \\
\cmidrule{1-6}\cmidrule{8-12}    SMD6  & \multicolumn{1}{{c}}{\cellcolor[rgb]{.851,.851,.851}1.00E-06 ($\approx$)} & \multicolumn{1}{{c}}{\cellcolor[rgb]{.851,.851,.851}1.00E-06 ($\approx$)} & \multicolumn{1}{{c}}{\cellcolor[rgb]{.851,.851,.851}4.71E+02 ($\approx$)} & \multicolumn{1}{{c}}{\cellcolor[rgb]{.851,.851,.851}1.63E+04 ($\approx$)} & \multicolumn{1}{{c}}{\cellcolor[rgb]{.851,.851,.851}1.68E+04 ($\approx$)} &       & \cellcolor[rgb]{.851,.851,.851}1.00E-06 & \cellcolor[rgb]{.851,.851,.851}1.00E-06 & \multicolumn{1}{c}{4.74E+02} & \multicolumn{1}{c}{1.64E+04} & 1.69E+04 (-0.6\%) \\
\cmidrule{1-6}\cmidrule{8-12}    SMD7  & \multicolumn{1}{{c}}{9.82E-02 (+)} & \multicolumn{1}{{c}}{2.44E+02 (+)} & \multicolumn{1}{{c}}{6.33E+02 (+)} & \multicolumn{1}{{c}}{1.93E+04 (+)} & \multicolumn{1}{{c}}{2.00E+04 (+)} &       & \cellcolor[rgb]{.851,.851,.851}1.00E-06 & \cellcolor[rgb]{.851,.851,.851}1.26E-06 & \multicolumn{1}{c}{\cellcolor[rgb]{.851,.851,.851}2.17E+02} & \multicolumn{1}{c}{\cellcolor[rgb]{.851,.851,.851}8.25E+03} & \cellcolor[rgb]{.851,.851,.851}8.47E+03 (57.6\%) \\
\cmidrule{1-6}\cmidrule{8-12}    SMD8  & \multicolumn{1}{{c}}{\cellcolor[rgb]{.851,.851,.851}1.00E-06 ($\approx$)} & \multicolumn{1}{{c}}{\cellcolor[rgb]{.851,.851,.851}1.00E-06 ($\approx$)} & \multicolumn{1}{{c}}{1.43E+03 (+)} & \multicolumn{1}{{c}}{4.72E+04 (+)} & \multicolumn{1}{{c}}{4.86E+04 (+)} &       & \cellcolor[rgb]{.851,.851,.851}1.00E-06 & \cellcolor[rgb]{.851,.851,.851}1.00E-06 & \multicolumn{1}{c}{\cellcolor[rgb]{.851,.851,.851}7.72E+02} & \multicolumn{1}{c}{\cellcolor[rgb]{.851,.851,.851}2.82E+04} & \cellcolor[rgb]{.851,.851,.851}2.89E+04 (40.4\%) \\
\cmidrule{1-6}\cmidrule{8-12}    SMD9  & \multicolumn{1}{{c}}{\cellcolor[rgb]{.851,.851,.851}1.00E-06 ($\approx$)} & \multicolumn{1}{{c}}{1.73E-06 ($\approx$)} & \multicolumn{1}{{c}}{3.52E+02 (+)} & \multicolumn{1}{{c}}{1.25E+04 (+)} & \multicolumn{1}{{c}}{1.29E+04 (+)} &       & \cellcolor[rgb]{.851,.851,.851}1.00E-06 & \cellcolor[rgb]{.851,.851,.851}1.07E-06 & \multicolumn{1}{c}{\cellcolor[rgb]{.851,.851,.851}2.29E+02} & \multicolumn{1}{c}{\cellcolor[rgb]{.851,.851,.851}9.09E+03} & \cellcolor[rgb]{.851,.851,.851}9.32E+03 (27.6\%) \\
\cmidrule{1-6}\cmidrule{8-12}    SMD10 & \multicolumn{1}{{c}}{1.60E+01 (+)} & \multicolumn{1}{{c}}{\cellcolor[rgb]{.851,.851,.851}1.00E-06 (-)} & \multicolumn{1}{{c}}{1.67E+03 (+)} & \multicolumn{1}{{c}}{5.41E+04 (+)} & \multicolumn{1}{{c}}{5.58E+04 (+)} &       & \cellcolor[rgb]{.851,.851,.851}4.62E-03 & 2.48E-03 & \multicolumn{1}{c}{\cellcolor[rgb]{.851,.851,.851}1.13E+03} & \multicolumn{1}{c}{\cellcolor[rgb]{.851,.851,.851}3.96E+04} & \cellcolor[rgb]{.851,.851,.851}4.07E+04 (26.9\%) \\
\cmidrule{1-6}\cmidrule{8-12}    SMD11 & \multicolumn{1}{{c}}{1.85E-03 (+)} & \multicolumn{1}{{c}}{2.97E-03 (+)} & \multicolumn{1}{{c}}{2.51E+03 (+)} & \multicolumn{1}{{c}}{8.29E+04 (+)} & \multicolumn{1}{{c}}{8.54E+04 (+)} &       & \cellcolor[rgb]{.851,.851,.851}1.00E-06 & \cellcolor[rgb]{.851,.851,.851}1.80E-04 & \multicolumn{1}{c}{\cellcolor[rgb]{.851,.851,.851}9.94E+02} & \multicolumn{1}{c}{\cellcolor[rgb]{.851,.851,.851}3.55E+04} & \cellcolor[rgb]{.851,.851,.851}3.65E+04 (57.2\%) \\
\cmidrule{1-6}\cmidrule{8-12}    SMD12 & \multicolumn{1}{{c}}{\cellcolor[rgb]{.851,.851,.851}1.00E-06 ($\approx$)} & \multicolumn{1}{{c}}{\cellcolor[rgb]{.851,.851,.851}1.60E+01 ($\approx$)} & \multicolumn{1}{{c}}{9.83E+02 (+)} & \multicolumn{1}{{c}}{3.24E+04 (+)} & \multicolumn{1}{{c}}{3.34E+04 (+)} &       & \cellcolor[rgb]{.851,.851,.851}1.00E-06 & \cellcolor[rgb]{.851,.851,.851}1.60E+01 & \multicolumn{1}{c}{\cellcolor[rgb]{.851,.851,.851}7.91E+02} & \multicolumn{1}{c}{\cellcolor[rgb]{.851,.851,.851}2.90E+04} & \cellcolor[rgb]{.851,.851,.851}2.98E+04 (10.9\%) \\
\cmidrule{1-6}\cmidrule{8-12}    +/$\approx$/- & 3/9/0 & 2/9/1 & 11/1/0 & 11/1/0 & 11/1/0 &       &       &       & \multicolumn{2}{r}{Average $R_{rs}$} & 33.1\% \\
    \bottomrule
    \end{tabular}%
}
\end{table*}%

\begin{table*}[htbp]
  \centering
  \caption{Performance comparison between BL-CMA-ES and CR-BL-CMA-ES regarding the median results of the accuracy and the number of FEs on SMD problems in 21 runs.}
    \label{tab:BLCMAES_vs_CRBLCMAES}
\resizebox{\textwidth}{!}{
    \begin{tabular}{ccccccccccrc}
    \toprule
    \multirow{2}[4]{*}{Probelm} & \multicolumn{5}{c}{BL-CMA-ES}         &       & \multicolumn{5}{c}{CR-BL-CMA-ES} \\
\cmidrule{2-6}\cmidrule{8-12}          & $Acc_u$  & $Acc_l$  & $FEs_u$  & $FEs_l$  & $FEs_t$  &       & $Acc_u$  & $Acc_l$  & \multicolumn{1}{c}{$FEs_u$} & \multicolumn{1}{c}{$FEs_l$} & $FEs_t$ ($R_{rs}$)\\
\cmidrule{1-6}\cmidrule{8-12}    SMD1  & \multicolumn{1}{{c}}{\cellcolor[rgb]{.851,.851,.851}1.00E-06 ($\approx$)} & \multicolumn{1}{{c}}{\cellcolor[rgb]{.851,.851,.851}1.00E-06 ($\approx$)} & \multicolumn{1}{{c}}{3.22E+02 (+)} & \multicolumn{1}{{c}}{2.08E+04 (+)} & \multicolumn{1}{{c}}{2.11E+04 (+)} &       & \cellcolor[rgb]{.851,.851,.851}1.00E-06 & \cellcolor[rgb]{.851,.851,.851}1.00E-06 & \multicolumn{1}{c}{\cellcolor[rgb]{.851,.851,.851}1.85E+02} & \multicolumn{1}{c}{\cellcolor[rgb]{.851,.851,.851}1.27E+04} & \cellcolor[rgb]{.851,.851,.851}1.29E+04 (38.8\%) \\
\cmidrule{1-6}\cmidrule{8-12}    SMD2  & \multicolumn{1}{{c}}{\cellcolor[rgb]{.851,.851,.851}1.00E-06 ($\approx$)} & \multicolumn{1}{{c}}{\cellcolor[rgb]{.851,.851,.851}1.00E-06 ($\approx$)} & \multicolumn{1}{{c}}{2.98E+02 (+)} & \multicolumn{1}{{c}}{2.02E+04 (+)} & \multicolumn{1}{{c}}{2.05E+04 (+)} &       & \cellcolor[rgb]{.851,.851,.851}1.00E-06 & 1.30E-06 & \multicolumn{1}{c}{\cellcolor[rgb]{.851,.851,.851}1.75E+02} & \multicolumn{1}{c}{\cellcolor[rgb]{.851,.851,.851}1.25E+04} & \cellcolor[rgb]{.851,.851,.851}1.27E+04 (37.8\%) \\
\cmidrule{1-6}\cmidrule{8-12}    SMD3  & \multicolumn{1}{{c}}{\cellcolor[rgb]{.851,.851,.851}1.00E-06 ($\approx$)} & \multicolumn{1}{{c}}{1.52E-06 (+)} & \multicolumn{1}{{c}}{3.13E+02 (+)} & \multicolumn{1}{{c}}{2.06E+04 (+)} & \multicolumn{1}{{c}}{2.09E+04 (+)} &       & \cellcolor[rgb]{.851,.851,.851}1.00E-06 & \cellcolor[rgb]{.851,.851,.851}1.00E-06 & \multicolumn{1}{c}{\cellcolor[rgb]{.851,.851,.851}1.78E+02} & \multicolumn{1}{c}{\cellcolor[rgb]{.851,.851,.851}1.32E+04} & \cellcolor[rgb]{.851,.851,.851}1.34E+04 (36.0\%) \\
\cmidrule{1-6}\cmidrule{8-12}    SMD4  & \multicolumn{1}{{c}}{\cellcolor[rgb]{.851,.851,.851}1.00E-06 ($\approx$)} & \multicolumn{1}{{c}}{3.39E-06 ($\approx$)} & \multicolumn{1}{{c}}{3.17E+02 (+)} & \multicolumn{1}{{c}}{2.13E+04 (+)} & \multicolumn{1}{{c}}{2.16E+04 (+)} &       & \cellcolor[rgb]{.851,.851,.851}1.00E-06 & \cellcolor[rgb]{.851,.851,.851}2.37E-06 & \multicolumn{1}{c}{\cellcolor[rgb]{.851,.851,.851}1.98E+02} & \multicolumn{1}{c}{\cellcolor[rgb]{.851,.851,.851}1.28E+04} & \cellcolor[rgb]{.851,.851,.851}1.30E+04 (39.9\%) \\
\cmidrule{1-6}\cmidrule{8-12}    SMD5  & \multicolumn{1}{{c}}{\cellcolor[rgb]{.851,.851,.851}1.00E-06 ($\approx$)} & \multicolumn{1}{{c}}{2.20E-06 ($\approx$)} & \multicolumn{1}{{c}}{3.68E+02 (+)} & \multicolumn{1}{{c}}{2.20E+04 (+)} & \multicolumn{1}{{c}}{2.24E+04 (+)} &       & \cellcolor[rgb]{.851,.851,.851}1.00E-06 & \cellcolor[rgb]{.851,.851,.851}1.00E-06 & \multicolumn{1}{c}{\cellcolor[rgb]{.851,.851,.851}2.22E+02} & \multicolumn{1}{c}{\cellcolor[rgb]{.851,.851,.851}1.36E+04} & \cellcolor[rgb]{.851,.851,.851}1.39E+04 (38.0\%) \\
\cmidrule{1-6}\cmidrule{8-12}    SMD6  & \multicolumn{1}{{c}}{\cellcolor[rgb]{.851,.851,.851}1.00E-06 ($\approx$)} & \multicolumn{1}{{c}}{\cellcolor[rgb]{.851,.851,.851}1.00E-06 ($\approx$)} & \multicolumn{1}{{c}}{4.30E+02 (+)} & \multicolumn{1}{{c}}{2.40E+04 (+)} & \multicolumn{1}{{c}}{2.45E+04 (+)} &       & \cellcolor[rgb]{.851,.851,.851}1.00E-06 & \cellcolor[rgb]{.851,.851,.851}1.00E-06 & \multicolumn{1}{c}{\cellcolor[rgb]{.851,.851,.851}3.95E+02} & \multicolumn{1}{c}{\cellcolor[rgb]{.851,.851,.851}2.27E+04} & \cellcolor[rgb]{.851,.851,.851}2.31E+04 (5.4\%) \\
\cmidrule{1-6}\cmidrule{8-12}    SMD7  & \multicolumn{1}{{c}}{\cellcolor[rgb]{.851,.851,.851}1.00E-06 ($\approx$)} & \multicolumn{1}{{c}}{\cellcolor[rgb]{.851,.851,.851}1.00E-06 ($\approx$)} & \multicolumn{1}{{c}}{3.36E+02 (+)} & \multicolumn{1}{{c}}{2.27E+04 (+)} & \multicolumn{1}{{c}}{2.31E+04 (+)} &       & \cellcolor[rgb]{.851,.851,.851}1.00E-06 & \cellcolor[rgb]{.851,.851,.851}1.00E-06 & \multicolumn{1}{c}{\cellcolor[rgb]{.851,.851,.851}1.87E+02} & \multicolumn{1}{c}{\cellcolor[rgb]{.851,.851,.851}1.38E+04} & \cellcolor[rgb]{.851,.851,.851}1.40E+04 (39.3\%) \\
\cmidrule{1-6}\cmidrule{8-12}    SMD8  & \multicolumn{1}{{c}}{\cellcolor[rgb]{.851,.851,.851}1.00E-06 ($\approx$)} & \multicolumn{1}{{c}}{\cellcolor[rgb]{.851,.851,.851}1.00E-06 ($\approx$)} & \multicolumn{1}{{c}}{1.35E+03 (+)} & \multicolumn{1}{{c}}{6.83E+04 (+)} & \multicolumn{1}{{c}}{6.97E+04 (+)} &       & \cellcolor[rgb]{.851,.851,.851}1.00E-06 & \cellcolor[rgb]{.851,.851,.851}1.00E-06 & \multicolumn{1}{c}{\cellcolor[rgb]{.851,.851,.851}7.63E+02} & \multicolumn{1}{c}{\cellcolor[rgb]{.851,.851,.851}3.83E+04} & \cellcolor[rgb]{.851,.851,.851}3.90E+04 (44.0\%) \\
\cmidrule{1-6}\cmidrule{8-12}    SMD9  & \multicolumn{1}{{c}}{\cellcolor[rgb]{.851,.851,.851}1.00E-06 ($\approx$)} & \multicolumn{1}{{c}}{\cellcolor[rgb]{.851,.851,.851}1.00E-06 ($\approx$)} & \multicolumn{1}{{c}}{3.02E+02 (+)} & \multicolumn{1}{{c}}{1.96E+04 (+)} & \multicolumn{1}{{c}}{1.99E+04 (+)} &       & \cellcolor[rgb]{.851,.851,.851}1.00E-06 & \cellcolor[rgb]{.851,.851,.851}1.00E-06 & \multicolumn{1}{c}{\cellcolor[rgb]{.851,.851,.851}2.26E+02} & \multicolumn{1}{c}{\cellcolor[rgb]{.851,.851,.851}1.41E+04} & \cellcolor[rgb]{.851,.851,.851}1.43E+04 (27.9\%) \\
\cmidrule{1-6}\cmidrule{8-12}    SMD10 & \multicolumn{1}{{c}}{1.60E+01 (+)} & \multicolumn{1}{{c}}{\cellcolor[rgb]{.851,.851,.851}1.00E-06 (-)} & \multicolumn{1}{{c}}{2.50E+03 (+)} & \multicolumn{1}{{c}}{9.30E+04 (+)} & \multicolumn{1}{{c}}{9.55E+04 (+)} &       & \cellcolor[rgb]{.851,.851,.851}1.15E-01 & 6.24E-02 & \multicolumn{1}{c}{\cellcolor[rgb]{.851,.851,.851}9.36E+02} & \multicolumn{1}{c}{\cellcolor[rgb]{.851,.851,.851}4.56E+04} & \cellcolor[rgb]{.851,.851,.851}4.65E+04 (51.3\%) \\
\cmidrule{1-6}\cmidrule{8-12}    SMD11 & \multicolumn{1}{{c}}{8.87E-04 (+)} & \multicolumn{1}{{c}}{9.74E-04 (+)} & \multicolumn{1}{{c}}{2.51E+03 (+)} & \multicolumn{1}{{c}}{1.23E+05 (+)} & \multicolumn{1}{{c}}{1.26E+05 (+)} &       & \cellcolor[rgb]{.851,.851,.851}1.00E-06 & \cellcolor[rgb]{.851,.851,.851}6.58E-05 & \multicolumn{1}{c}{\cellcolor[rgb]{.851,.851,.851}1.66E+03} & \multicolumn{1}{c}{\cellcolor[rgb]{.851,.851,.851}7.90E+04} & \cellcolor[rgb]{.851,.851,.851}8.06E+04 (35.8\%) \\
\cmidrule{1-6}\cmidrule{8-12}    SMD12 & \multicolumn{1}{{c}}{\cellcolor[rgb]{.851,.851,.851}1.00E-06 ($\approx$)} & \multicolumn{1}{{c}}{6.28E-05 (+)} & \multicolumn{1}{{c}}{9.43E+02 (+)} & \multicolumn{1}{{c}}{4.64E+04 (+)} & \multicolumn{1}{{c}}{4.73E+04 (+)} &       & \cellcolor[rgb]{.851,.851,.851}1.00E-06 & \cellcolor[rgb]{.851,.851,.851}4.77E-06 & \multicolumn{1}{c}{\cellcolor[rgb]{.851,.851,.851}7.35E+02} & \multicolumn{1}{c}{\cellcolor[rgb]{.851,.851,.851}4.02E+04} & \cellcolor[rgb]{.851,.851,.851}4.10E+04 (13.5\%) \\
\cmidrule{1-6}\cmidrule{8-12}    +/$\approx$/- & 2/10/0 & 3/8/1 & 12/0/0 & 12/0/0 & 12/0/0 &       &       &       & \multicolumn{2}{r}{Average $R_{rs}$} & 34.0\% \\
    \bottomrule
    \end{tabular}%
}
\end{table*}%

\begin{table*}[htbp]
  \centering
  \caption{Performance comparison between MOTEA and CR-MOTEA regarding the median results of the accuracy and the number of FEs on SMD problems in 21 runs.}
    \label{tab:MOTEA_vs_CRMOTEA}
\resizebox{\textwidth}{!}{
    \begin{tabular}{ccccccccccrc}
    \toprule
    \multirow{2}[4]{*}{Probelm} & \multicolumn{5}{c}{MOTEA}             &       & \multicolumn{5}{c}{CR-MOTEA} \\
\cmidrule{2-6}\cmidrule{8-12}          & $Acc_u$  & $Acc_l$  & $FEs_u$  & $FEs_l$  & $FEs_t$  &       & $Acc_u$  & $Acc_l$  & \multicolumn{1}{c}{$FEs_u$} & \multicolumn{1}{c}{$FEs_l$} & $FEs_t$ ($R_{rs}$)\\
\cmidrule{1-6}\cmidrule{8-12}    SMD1  & \multicolumn{1}{{c}}{\cellcolor[rgb]{.851,.851,.851}1.00E-06 ($\approx$)} & \multicolumn{1}{{c}}{\cellcolor[rgb]{.851,.851,.851}1.00E-06 ($\approx$)} & \multicolumn{1}{{c}}{3.09E+02 (+)} & \multicolumn{1}{{c}}{2.01E+04 (+)} & \multicolumn{1}{{c}}{2.04E+04 (+)} &       & \cellcolor[rgb]{.851,.851,.851}1.00E-06 & \cellcolor[rgb]{.851,.851,.851}1.00E-06 & \multicolumn{1}{c}{\cellcolor[rgb]{.851,.851,.851}1.91E+02} & \multicolumn{1}{c}{\cellcolor[rgb]{.851,.851,.851}1.12E+04} & \cellcolor[rgb]{.851,.851,.851}1.14E+04 (44.1\%) \\
\cmidrule{1-6}\cmidrule{8-12}    SMD2  & \multicolumn{1}{{c}}{\cellcolor[rgb]{.851,.851,.851}1.00E-06 ($\approx$)} & \multicolumn{1}{{c}}{1.19E-06 ($\approx$)} & \multicolumn{1}{{c}}{2.85E+02 (+)} & \multicolumn{1}{{c}}{2.01E+04 (+)} & \multicolumn{1}{{c}}{2.03E+04 (+)} &       & \cellcolor[rgb]{.851,.851,.851}1.00E-06 & \cellcolor[rgb]{.851,.851,.851}1.00E-06 & \multicolumn{1}{c}{\cellcolor[rgb]{.851,.851,.851}1.81E+02} & \multicolumn{1}{c}{\cellcolor[rgb]{.851,.851,.851}1.04E+04} & \cellcolor[rgb]{.851,.851,.851}1.06E+04 (48.1\%) \\
\cmidrule{1-6}\cmidrule{8-12}    SMD3  & \multicolumn{1}{{c}}{\cellcolor[rgb]{.851,.851,.851}1.00E-06 ($\approx$)} & \multicolumn{1}{{c}}{\cellcolor[rgb]{.851,.851,.851}1.00E-06 ($\approx$)} & \multicolumn{1}{{c}}{3.15E+02 (+)} & \multicolumn{1}{{c}}{1.69E+04 (+)} & \multicolumn{1}{{c}}{1.72E+04 (+)} &       & \cellcolor[rgb]{.851,.851,.851}1.00E-06 & \cellcolor[rgb]{.851,.851,.851}1.00E-06 & \multicolumn{1}{c}{\cellcolor[rgb]{.851,.851,.851}1.65E+02} & \multicolumn{1}{c}{\cellcolor[rgb]{.851,.851,.851}9.79E+03} & \cellcolor[rgb]{.851,.851,.851}9.95E+03 (42.2\%) \\
\cmidrule{1-6}\cmidrule{8-12}    SMD4  & \multicolumn{1}{{c}}{\cellcolor[rgb]{.851,.851,.851}1.00E-06 ($\approx$)} & \multicolumn{1}{{c}}{3.84E-06 (+)} & \multicolumn{1}{{c}}{2.96E+02 (+)} & \multicolumn{1}{{c}}{1.67E+04 (+)} & \multicolumn{1}{{c}}{1.70E+04 (+)} &       & \cellcolor[rgb]{.851,.851,.851}1.00E-06 & \cellcolor[rgb]{.851,.851,.851}1.08E-06 & \multicolumn{1}{c}{\cellcolor[rgb]{.851,.851,.851}1.92E+02} & \multicolumn{1}{c}{\cellcolor[rgb]{.851,.851,.851}9.06E+03} & \cellcolor[rgb]{.851,.851,.851}9.25E+03 (45.6\%) \\
\cmidrule{1-6}\cmidrule{8-12}    SMD5  & \multicolumn{1}{{c}}{\cellcolor[rgb]{.851,.851,.851}1.00E-06 ($\approx$)} & \multicolumn{1}{{c}}{1.81E-06 ($\approx$)} & \multicolumn{1}{{c}}{3.09E+02 (+)} & \multicolumn{1}{{c}}{1.83E+04 (+)} & \multicolumn{1}{{c}}{1.86E+04 (+)} &       & \cellcolor[rgb]{.851,.851,.851}1.00E-06 & \cellcolor[rgb]{.851,.851,.851}1.20E-06 & \multicolumn{1}{c}{\cellcolor[rgb]{.851,.851,.851}2.21E+02} & \multicolumn{1}{c}{\cellcolor[rgb]{.851,.851,.851}1.08E+04} & \cellcolor[rgb]{.851,.851,.851}1.10E+04 (40.7\%) \\
\cmidrule{1-6}\cmidrule{8-12}    SMD6  & \multicolumn{1}{{c}}{\cellcolor[rgb]{.851,.851,.851}1.00E-06 ($\approx$)} & \multicolumn{1}{{c}}{\cellcolor[rgb]{.851,.851,.851}1.00E-06 ($\approx$)} & \multicolumn{1}{{c}}{\cellcolor[rgb]{.851,.851,.851}4.76E+02 (-)} & \multicolumn{1}{{c}}{2.11E+04 (+)} & \multicolumn{1}{{c}}{2.15E+04 (+)} &       & \cellcolor[rgb]{.851,.851,.851}1.00E-06 & \cellcolor[rgb]{.851,.851,.851}1.00E-06 & \multicolumn{1}{c}{5.42E+02} & \multicolumn{1}{c}{\cellcolor[rgb]{.851,.851,.851}1.86E+04} & \cellcolor[rgb]{.851,.851,.851}1.91E+04 (11.2\%) \\
\cmidrule{1-6}\cmidrule{8-12}    SMD7  & \multicolumn{1}{{c}}{3.20E-04 (+)} & \multicolumn{1}{{c}}{9.07E-06 (+)} & \multicolumn{1}{{c}}{7.57E+02 (+)} & \multicolumn{1}{{c}}{2.04E+04 (+)} & \multicolumn{1}{{c}}{2.12E+04 (+)} &       & \cellcolor[rgb]{.851,.851,.851}1.00E-06 & \cellcolor[rgb]{.851,.851,.851}1.00E-06 & \multicolumn{1}{c}{\cellcolor[rgb]{.851,.851,.851}1.82E+02} & \multicolumn{1}{c}{\cellcolor[rgb]{.851,.851,.851}1.14E+04} & \cellcolor[rgb]{.851,.851,.851}1.16E+04 (45.3\%) \\
\cmidrule{1-6}\cmidrule{8-12}    SMD8  & \multicolumn{1}{{c}}{\cellcolor[rgb]{.851,.851,.851}1.00E-06 ($\approx$)} & \multicolumn{1}{{c}}{\cellcolor[rgb]{.851,.851,.851}1.00E-06 ($\approx$)} & \multicolumn{1}{{c}}{1.38E+03 (+)} & \multicolumn{1}{{c}}{4.48E+04 (+)} & \multicolumn{1}{{c}}{4.62E+04 (+)} &       & \cellcolor[rgb]{.851,.851,.851}1.00E-06 & \cellcolor[rgb]{.851,.851,.851}1.00E-06 & \multicolumn{1}{c}{\cellcolor[rgb]{.851,.851,.851}6.90E+02} & \multicolumn{1}{c}{\cellcolor[rgb]{.851,.851,.851}2.35E+04} & \cellcolor[rgb]{.851,.851,.851}2.42E+04 (47.7\%) \\
\cmidrule{1-6}\cmidrule{8-12}    SMD9  & \multicolumn{1}{{c}}{\cellcolor[rgb]{.851,.851,.851}1.00E-06 ($\approx$)} & \multicolumn{1}{{c}}{\cellcolor[rgb]{.851,.851,.851}1.05E-06 ($\approx$)} & \multicolumn{1}{{c}}{3.00E+02 (+)} & \multicolumn{1}{{c}}{2.45E+04 (+)} & \multicolumn{1}{{c}}{2.48E+04 (+)} &       & \cellcolor[rgb]{.851,.851,.851}1.00E-06 & 1.05E-06 & \multicolumn{1}{c}{\cellcolor[rgb]{.851,.851,.851}2.38E+02} & \multicolumn{1}{c}{\cellcolor[rgb]{.851,.851,.851}1.35E+04} & \cellcolor[rgb]{.851,.851,.851}1.37E+04 (44.8\%) \\
\cmidrule{1-6}\cmidrule{8-12}    SMD10 & \multicolumn{1}{{c}}{\cellcolor[rgb]{.851,.851,.851}1.60E+01 ($\approx$)} & \multicolumn{1}{{c}}{\cellcolor[rgb]{.851,.851,.851}3.74E-06 (-)} & \multicolumn{1}{{c}}{1.31E+03 (+)} & \multicolumn{1}{{c}}{4.36E+04 (+)} & \multicolumn{1}{{c}}{4.49E+04 (+)} &       & \cellcolor[rgb]{.851,.851,.851}1.60E+01 & 1.36E-05 & \multicolumn{1}{c}{\cellcolor[rgb]{.851,.851,.851}9.51E+02} & \multicolumn{1}{c}{\cellcolor[rgb]{.851,.851,.851}3.10E+04} & \cellcolor[rgb]{.851,.851,.851}3.20E+04 (28.7\%) \\
\cmidrule{1-6}\cmidrule{8-12}    SMD11 & \multicolumn{1}{{c}}{2.03E-03 (+)} & \multicolumn{1}{{c}}{2.07E-03 (+)} & \multicolumn{1}{{c}}{2.50E+03 (+)} & \multicolumn{1}{{c}}{7.49E+04 (+)} & \multicolumn{1}{{c}}{7.74E+04 (+)} &       & \cellcolor[rgb]{.851,.851,.851}1.24E-04 & \cellcolor[rgb]{.851,.851,.851}2.89E-04 & \multicolumn{1}{c}{\cellcolor[rgb]{.851,.851,.851}2.16E+03} & \multicolumn{1}{c}{\cellcolor[rgb]{.851,.851,.851}5.82E+04} & \cellcolor[rgb]{.851,.851,.851}6.04E+04 (22.1\%) \\
\cmidrule{1-6}\cmidrule{8-12}    SMD12 & \multicolumn{1}{{c}}{\cellcolor[rgb]{.851,.851,.851}1.00E-06 ($\approx$)} & \multicolumn{1}{{c}}{5.05E-06 (+)} & \multicolumn{1}{{c}}{\cellcolor[rgb]{.851,.851,.851}8.35E+02 (-)} & \multicolumn{1}{{c}}{3.48E+04 (+)} & \multicolumn{1}{{c}}{3.57E+04 (+)} &       & \cellcolor[rgb]{.851,.851,.851}1.00E-06 & \cellcolor[rgb]{.851,.851,.851}2.61E-06 & \multicolumn{1}{c}{9.39E+02} & \multicolumn{1}{c}{\cellcolor[rgb]{.851,.851,.851}2.95E+04} & \cellcolor[rgb]{.851,.851,.851}3.04E+04 (14.8\%) \\
\cmidrule{1-6}\cmidrule{8-12}    +/$\approx$/- & 2/10/0 & 4/7/1 & 10/0/2 & 12/0/0 & 12/0/0 &       &       &       & \multicolumn{2}{r}{Average $R_{rs}$} & 36.3\% \\
    \bottomrule
    \end{tabular}%
}
\end{table*}%

\begin{table*}[htbp]
  \centering
  \caption{Performance comparison between BOC and CR-BOC regarding the median results of the accuracy and the number of FEs on SMD problems in 21 runs.}
   \label{tab:BOC_vs_CRBOC}
\resizebox{\textwidth}{!}{
    \begin{tabular}{ccccccccccrc}
    \toprule
    \multirow{2}[4]{*}{Probelm} & \multicolumn{5}{c}{BOC}               &       & \multicolumn{5}{c}{CR-BOC} \\
\cmidrule{2-6}\cmidrule{8-12}          & $Acc_u$  & $Acc_l$  & $FEs_u$  & $FEs_l$  & $FEs_t$  &       & $Acc_u$  & $Acc_l$  & \multicolumn{1}{c}{$FEs_u$} & \multicolumn{1}{c}{$FEs_l$} & $FEs_t$ ($R_{rs}$)\\
\cmidrule{1-6}\cmidrule{8-12}    SMD1  & \multicolumn{1}{{c}}{\cellcolor[rgb]{.851,.851,.851}1.00E-06 ($\approx$)} & \multicolumn{1}{{c}}{\cellcolor[rgb]{.851,.851,.851}1.00E-06 ($\approx$)} & \multicolumn{1}{{c}}{3.36E+02 (+)} & \multicolumn{1}{{c}}{1.09E+04 (+)} & \multicolumn{1}{{c}}{1.12E+04 (+)} &       & \cellcolor[rgb]{.851,.851,.851}1.00E-06 & \cellcolor[rgb]{.851,.851,.851}1.00E-06 & \multicolumn{1}{c}{\cellcolor[rgb]{.851,.851,.851}1.96E+02} & \multicolumn{1}{c}{\cellcolor[rgb]{.851,.851,.851}6.14E+03} & \cellcolor[rgb]{.851,.851,.851}6.33E+03 (43.4\%) \\
\cmidrule{1-6}\cmidrule{8-12}    SMD2  & \multicolumn{1}{{c}}{\cellcolor[rgb]{.851,.851,.851}1.00E-06 ($\approx$)} & \multicolumn{1}{{c}}{\cellcolor[rgb]{.851,.851,.851}1.25E-06 ($\approx$)} & \multicolumn{1}{{c}}{2.89E+02 (+)} & \multicolumn{1}{{c}}{9.89E+03 (+)} & \multicolumn{1}{{c}}{1.02E+04 (+)} &       & \cellcolor[rgb]{.851,.851,.851}1.00E-06 & 1.27E-06 & \multicolumn{1}{c}{\cellcolor[rgb]{.851,.851,.851}1.73E+02} & \multicolumn{1}{c}{\cellcolor[rgb]{.851,.851,.851}5.21E+03} & \cellcolor[rgb]{.851,.851,.851}5.38E+03 (47.1\%) \\
\cmidrule{1-6}\cmidrule{8-12}    SMD3  & \multicolumn{1}{{c}}{\cellcolor[rgb]{.851,.851,.851}1.00E-06 ($\approx$)} & \multicolumn{1}{{c}}{\cellcolor[rgb]{.851,.851,.851}1.00E-06 ($\approx$)} & \multicolumn{1}{{c}}{3.26E+02 (+)} & \multicolumn{1}{{c}}{9.62E+03 (+)} & \multicolumn{1}{{c}}{9.95E+03 (+)} &       & \cellcolor[rgb]{.851,.851,.851}1.00E-06 & \cellcolor[rgb]{.851,.851,.851}1.00E-06 & \multicolumn{1}{c}{\cellcolor[rgb]{.851,.851,.851}2.15E+02} & \multicolumn{1}{c}{\cellcolor[rgb]{.851,.851,.851}6.23E+03} & \cellcolor[rgb]{.851,.851,.851}6.44E+03 (35.2\%) \\
\cmidrule{1-6}\cmidrule{8-12}    SMD4  & \multicolumn{1}{{c}}{\cellcolor[rgb]{.851,.851,.851}1.00E-06 ($\approx$)} & \multicolumn{1}{{c}}{3.20E-06 ($\approx$)} & \multicolumn{1}{{c}}{3.25E+02 (+)} & \multicolumn{1}{{c}}{9.03E+03 (+)} & \multicolumn{1}{{c}}{9.36E+03 (+)} &       & \cellcolor[rgb]{.851,.851,.851}1.00E-06 & \cellcolor[rgb]{.851,.851,.851}1.36E-06 & \multicolumn{1}{c}{\cellcolor[rgb]{.851,.851,.851}2.24E+02} & \multicolumn{1}{c}{\cellcolor[rgb]{.851,.851,.851}6.37E+03} & \cellcolor[rgb]{.851,.851,.851}6.60E+03 (29.5\%) \\
\cmidrule{1-6}\cmidrule{8-12}    SMD5  & \multicolumn{1}{{c}}{\cellcolor[rgb]{.851,.851,.851}1.00E-06 ($\approx$)} & \multicolumn{1}{{c}}{5.16E-06 (+)} & \multicolumn{1}{{c}}{3.56E+02 (+)} & \multicolumn{1}{{c}}{1.06E+04 (+)} & \multicolumn{1}{{c}}{1.10E+04 (+)} &       & \cellcolor[rgb]{.851,.851,.851}1.00E-06 & \cellcolor[rgb]{.851,.851,.851}1.38E-06 & \multicolumn{1}{c}{\cellcolor[rgb]{.851,.851,.851}2.43E+02} & \multicolumn{1}{c}{\cellcolor[rgb]{.851,.851,.851}6.49E+03} & \cellcolor[rgb]{.851,.851,.851}6.73E+03 (38.8\%) \\
\cmidrule{1-6}\cmidrule{8-12}    SMD6  & \multicolumn{1}{{c}}{\cellcolor[rgb]{.851,.851,.851}1.00E-06 (-)} & \multicolumn{1}{{c}}{\cellcolor[rgb]{.851,.851,.851}1.00E-06 (-)} & \multicolumn{1}{{c}}{\cellcolor[rgb]{.851,.851,.851}6.16E+02 ($\approx$)} & \multicolumn{1}{{c}}{\cellcolor[rgb]{.851,.851,.851}1.95E+04 ($\approx$)} & \multicolumn{1}{{c}}{\cellcolor[rgb]{.851,.851,.851}2.02E+04 ($\approx$)} &       & 3.37E-01 & 1.17E-01 & \multicolumn{1}{c}{6.55E+02} & \multicolumn{1}{c}{1.96E+04} & 2.03E+04 (-0.5\%) \\
\cmidrule{1-6}\cmidrule{8-12}    SMD7  & \multicolumn{1}{{c}}{\cellcolor[rgb]{.851,.851,.851}1.00E-06 ($\approx$)} & \multicolumn{1}{{c}}{\cellcolor[rgb]{.851,.851,.851}1.00E-06 ($\approx$)} & \multicolumn{1}{{c}}{3.34E+02 (+)} & \multicolumn{1}{{c}}{1.10E+04 (+)} & \multicolumn{1}{{c}}{1.13E+04 (+)} &       & \cellcolor[rgb]{.851,.851,.851}1.00E-06 & \cellcolor[rgb]{.851,.851,.851}1.00E-06 & \multicolumn{1}{c}{\cellcolor[rgb]{.851,.851,.851}2.47E+02} & \multicolumn{1}{c}{\cellcolor[rgb]{.851,.851,.851}7.36E+03} & \cellcolor[rgb]{.851,.851,.851}7.61E+03 (32.8\%) \\
\cmidrule{1-6}\cmidrule{8-12}    SMD8  & \multicolumn{1}{{c}}{\cellcolor[rgb]{.851,.851,.851}1.00E-06 ($\approx$)} & \multicolumn{1}{{c}}{\cellcolor[rgb]{.851,.851,.851}1.00E-06 ($\approx$)} & \multicolumn{1}{{c}}{1.41E+03 (+)} & \multicolumn{1}{{c}}{4.34E+04 (+)} & \multicolumn{1}{{c}}{4.48E+04 (+)} &       & \cellcolor[rgb]{.851,.851,.851}1.00E-06 & \cellcolor[rgb]{.851,.851,.851}1.00E-06 & \multicolumn{1}{c}{\cellcolor[rgb]{.851,.851,.851}6.79E+02} & \multicolumn{1}{c}{\cellcolor[rgb]{.851,.851,.851}1.98E+04} & \cellcolor[rgb]{.851,.851,.851}2.04E+04 (54.4\%) \\
\cmidrule{1-6}\cmidrule{8-12}    SMD9  & \multicolumn{1}{{c}}{\cellcolor[rgb]{.851,.851,.851}1.00E-06 ($\approx$)} & \multicolumn{1}{{c}}{\cellcolor[rgb]{.851,.851,.851}1.00E-06 ($\approx$)} & \multicolumn{1}{{c}}{3.02E+02 (+)} & \multicolumn{1}{{c}}{9.93E+03 (+)} & \multicolumn{1}{{c}}{1.02E+04 (+)} &       & \cellcolor[rgb]{.851,.851,.851}1.00E-06 & \cellcolor[rgb]{.851,.851,.851}1.00E-06 & \multicolumn{1}{c}{\cellcolor[rgb]{.851,.851,.851}2.13E+02} & \multicolumn{1}{c}{\cellcolor[rgb]{.851,.851,.851}6.29E+03} & \cellcolor[rgb]{.851,.851,.851}6.50E+03 (36.4\%) \\
\cmidrule{1-6}\cmidrule{8-12}    SMD10 & \multicolumn{1}{{c}}{1.60E+01 (+)} & \multicolumn{1}{{c}}{1.60E+01 (+)} & \multicolumn{1}{{c}}{1.79E+03 (+)} & \multicolumn{1}{{c}}{4.64E+04 (+)} & \multicolumn{1}{{c}}{4.82E+04 (+)} &       & \cellcolor[rgb]{.851,.851,.851}2.71E+00 & \cellcolor[rgb]{.851,.851,.851}8.02E-01 & \multicolumn{1}{c}{\cellcolor[rgb]{.851,.851,.851}1.18E+03} & \multicolumn{1}{c}{\cellcolor[rgb]{.851,.851,.851}3.05E+04} & \cellcolor[rgb]{.851,.851,.851}3.16E+04 (34.4\%) \\
\cmidrule{1-6}\cmidrule{8-12}    SMD11 & \multicolumn{1}{{c}}{8.88E-03 (+)} & \multicolumn{1}{{c}}{5.00E-02 (+)} & \multicolumn{1}{{c}}{2.51E+03 (+)} & \multicolumn{1}{{c}}{7.23E+04 (+)} & \multicolumn{1}{{c}}{7.49E+04 (+)} &       & \cellcolor[rgb]{.851,.851,.851}2.23E-03 & \cellcolor[rgb]{.851,.851,.851}2.90E-03 & \multicolumn{1}{c}{\cellcolor[rgb]{.851,.851,.851}2.50E+03} & \multicolumn{1}{c}{\cellcolor[rgb]{.851,.851,.851}6.30E+04} & \cellcolor[rgb]{.851,.851,.851}6.55E+04 (12.5\%) \\
\cmidrule{1-6}\cmidrule{8-12}    SMD12 & \multicolumn{1}{{c}}{\cellcolor[rgb]{.851,.851,.851}1.00E-06 ($\approx$)} & \multicolumn{1}{{c}}{\cellcolor[rgb]{.851,.851,.851}1.60E+01 ($\approx$)} & \multicolumn{1}{{c}}{9.61E+02 (+)} & \multicolumn{1}{{c}}{2.63E+04 (+)} & \multicolumn{1}{{c}}{2.72E+04 (+)} &       & \cellcolor[rgb]{.851,.851,.851}1.00E-06 & \cellcolor[rgb]{.851,.851,.851}1.60E+01 & \multicolumn{1}{c}{\cellcolor[rgb]{.851,.851,.851}7.09E+02} & \multicolumn{1}{c}{\cellcolor[rgb]{.851,.851,.851}1.94E+04} & \cellcolor[rgb]{.851,.851,.851}2.01E+04 (26.3\%) \\
\cmidrule{1-6}\cmidrule{8-12}    +/$\approx$/- & 2/9/1 & 3/8/1 & 11/1/0 & 11/1/0 & 11/1/0 &       &       &       & \multicolumn{2}{r}{Average $R_{rs}$} & 32.5\% \\
    \bottomrule
    \end{tabular}%
}
\end{table*}%

As can be observed form Tables \ref{tab:TLEACMAES_vs_CRTLEACMAES} - \ref{tab:BOC_vs_CRBOC}, 
the variants incorporated with CR-BLEA achieve equivalent or better accuracy than the original algorithms on almost all the SMD problems.
Taking Table \ref{tab:TLEACMAES_vs_CRTLEACMAES} as an example, the statistical results regarding accuracy indicate that there is no significant difference between TLEA-CMA-ES and CR-TLEA-CMA-ES in 10 out of 12 results.
Although CR-BLEA carries the risk of accuracy deterioration because it suggests not to optimize all lower-level tasks.
The data-driven contrastive ranking model effectively helps the algorithm identify promising parts, which in turn guarantees the accuracy.
For SMD10 and SMD11, the variant algorithms achiev significantly better accuracy than the original algorithms.
Because the original algorithms remain stuck in a local optimum with the limited 2500 upper-level FEs $FEs_u^{max}$,
while CR-BLEA converges before reaching the maximum upper-level FEs limit.

In terms of the number of function evaluations, the statistical results show that the variants consume significantly fewer upper- and lower-level FEs on all SMD problems except SMD6.
Compared to the original algorithms, since some unpromising tasks are not assigned execution opportunities, the resource wasted on them is saved, resulting in a substantial reduction of upper- and lower-level FEs consumption, especially at the lower level.
The reduced upper- and lower-level FEs also result in a lower total FEs consumption.
As observed from the resource-saving
rate regarding the $FEs_t$ reduction shown in parentheses, except for SMD6 and SMD12, the variants reduce total resource consumption by more than 30\% on average compared to the original algorithm across various problems, with a maximum reduction of up to 57.6\%.

\begin{table*}[htbp]
  \centering
  \caption{Performance comparison between TLEA-CMA-ES and CR-TLEA-CMA-ES regarding the median results of the accuracy and the number of FEs on TP problems in 21 runs.}
  \label{tab:TP_TLEACMAES_vs_CRTLEACMAES}
\resizebox{\textwidth}{!}{    
\begin{tabular}{ccccccccccrc}
    \toprule
    \multirow{2}[4]{*}{Probelm} & \multicolumn{5}{c}{TLEA-CMA-ES}       &       & \multicolumn{5}{c}{CR-TLEA-CMA-ES} \\
\cmidrule{2-6}\cmidrule{8-12}          & $Acc_u$  & $Acc_l$  & $FEs_u$  & $FEs_l$  & $FEs_t$  &       & $Acc_u$  & $Acc_l$  & \multicolumn{1}{c}{$FEs_u$} & \multicolumn{1}{c}{$FEs_l$} & $FEs_t$ ($R_{rs}$)\\
\cmidrule{1-6}\cmidrule{8-12}    TP1   & \multicolumn{1}{{c}}{\cellcolor[rgb]{.851,.851,.851}1.00E-06 ($\approx$)} & \multicolumn{1}{{c}}{4.12E-06 ($\approx$)} & \multicolumn{1}{{c}}{1.87E+03 (+)} & \multicolumn{1}{{c}}{8.08E+04 (+)} & \multicolumn{1}{{c}}{8.27E+04 (+)} &       & \cellcolor[rgb]{.851,.851,.851}1.00E-06 & \cellcolor[rgb]{.851,.851,.851}1.68E-06 & \multicolumn{1}{c}{\cellcolor[rgb]{.851,.851,.851}1.40E+03} & \multicolumn{1}{c}{\cellcolor[rgb]{.851,.851,.851}6.83E+04} & \cellcolor[rgb]{.851,.851,.851}6.97E+04 (15.7\%) \\
\cmidrule{1-6}\cmidrule{8-12}    TP2   & \multicolumn{1}{{c}}{1.33E-05 (+)} & \multicolumn{1}{{c}}{1.43E-04 (+)} & \multicolumn{1}{{c}}{1.85E+03 (+)} & \multicolumn{1}{{c}}{7.58E+04 (+)} & \multicolumn{1}{{c}}{7.77E+04 (+)} &       & \cellcolor[rgb]{.851,.851,.851}1.00E-06 & \cellcolor[rgb]{.851,.851,.851}1.09E-05 & \multicolumn{1}{c}{\cellcolor[rgb]{.851,.851,.851}9.84E+02} & \multicolumn{1}{c}{\cellcolor[rgb]{.851,.851,.851}4.43E+04} & \cellcolor[rgb]{.851,.851,.851}4.53E+04 (41.7\%) \\
\cmidrule{1-6}\cmidrule{8-12}    TP3   & \multicolumn{1}{{c}}{3.39E-02 ($\approx$)} & \multicolumn{1}{{c}}{\cellcolor[rgb]{.851,.851,.851}4.14E-04 ($\approx$)} & \multicolumn{1}{{c}}{2.51E+03 (+)} & \multicolumn{1}{{c}}{1.07E+05 (+)} & \multicolumn{1}{{c}}{1.09E+05 (+)} &       & \cellcolor[rgb]{.851,.851,.851}1.86E-02 & 7.22E-03 & \multicolumn{1}{c}{\cellcolor[rgb]{.851,.851,.851}9.27E+02} & \multicolumn{1}{c}{\cellcolor[rgb]{.851,.851,.851}6.05E+04} & \cellcolor[rgb]{.851,.851,.851}6.14E+04 (43.8\%) \\
\cmidrule{1-6}\cmidrule{8-12}    TP4   & \multicolumn{1}{{c}}{\cellcolor[rgb]{.851,.851,.851}1.00E-06 ($\approx$)} & \multicolumn{1}{{c}}{\cellcolor[rgb]{.851,.851,.851}1.00E-06 ($\approx$)} & \multicolumn{1}{{c}}{1.24E+03 (+)} & \multicolumn{1}{{c}}{5.51E+04 (+)} & \multicolumn{1}{{c}}{5.63E+04 (+)} &       & \cellcolor[rgb]{.851,.851,.851}1.00E-06 & \cellcolor[rgb]{.851,.851,.851}1.00E-06 & \multicolumn{1}{c}{\cellcolor[rgb]{.851,.851,.851}8.89E+02} & \multicolumn{1}{c}{\cellcolor[rgb]{.851,.851,.851}4.60E+04} & \cellcolor[rgb]{.851,.851,.851}4.69E+04 (16.8\%) \\
\cmidrule{1-6}\cmidrule{8-12}    TP5   & \multicolumn{1}{{c}}{4.23E-01 (+)} & \multicolumn{1}{{c}}{9.02E-02 (+)} & \multicolumn{1}{{c}}{1.76E+03 (+)} & \multicolumn{1}{{c}}{7.26E+04 (+)} & \multicolumn{1}{{c}}{7.44E+04 (+)} &       & \cellcolor[rgb]{.851,.851,.851}1.34E-04 & \cellcolor[rgb]{.851,.851,.851}8.53E-04 & \multicolumn{1}{c}{\cellcolor[rgb]{.851,.851,.851}7.23E+02} & \multicolumn{1}{c}{\cellcolor[rgb]{.851,.851,.851}4.06E+04} & \cellcolor[rgb]{.851,.851,.851}4.13E+04 (44.4\%) \\
\cmidrule{1-6}\cmidrule{8-12}    TP6   & \multicolumn{1}{{c}}{\cellcolor[rgb]{.851,.851,.851}1.00E-06 ($\approx$)} & \multicolumn{1}{{c}}{2.53E-05 ($\approx$)} & \multicolumn{1}{{c}}{5.46E+02 (+)} & \multicolumn{1}{{c}}{2.87E+04 (+)} & \multicolumn{1}{{c}}{2.92E+04 (+)} &       & \cellcolor[rgb]{.851,.851,.851}1.00E-06 & \cellcolor[rgb]{.851,.851,.851}2.38E-05 & \multicolumn{1}{c}{\cellcolor[rgb]{.851,.851,.851}5.00E+02} & \multicolumn{1}{c}{\cellcolor[rgb]{.851,.851,.851}2.66E+04} & \cellcolor[rgb]{.851,.851,.851}2.71E+04 (7.3\%) \\
\cmidrule{1-6}\cmidrule{8-12}    TP7   & \multicolumn{1}{{c}}{\cellcolor[rgb]{.851,.851,.851}6.41E-04 ($\approx$)} & \multicolumn{1}{{c}}{\cellcolor[rgb]{.851,.851,.851}6.41E-04 ($\approx$)} & \multicolumn{1}{{c}}{1.61E+03 ($\approx$)} & \multicolumn{1}{{c}}{7.96E+04 ($\approx$)} & \multicolumn{1}{{c}}{8.12E+04 ($\approx$)} &       & 6.66E-04 & 6.66E-04 & \multicolumn{1}{c}{\cellcolor[rgb]{.851,.851,.851}1.59E+03} & \multicolumn{1}{c}{\cellcolor[rgb]{.851,.851,.851}7.93E+04} & \cellcolor[rgb]{.851,.851,.851}8.09E+04 (0.4\%) \\
\cmidrule{1-6}\cmidrule{8-12}    TP8   & \multicolumn{1}{{c}}{\cellcolor[rgb]{.851,.851,.851}1.00E-06 ($\approx$)} & \multicolumn{1}{{c}}{3.86E-05 ($\approx$)} & \multicolumn{1}{{c}}{1.85E+03 (+)} & \multicolumn{1}{{c}}{7.61E+04 (+)} & \multicolumn{1}{{c}}{7.79E+04 (+)} &       & \cellcolor[rgb]{.851,.851,.851}1.00E-06 & \cellcolor[rgb]{.851,.851,.851}5.11E-06 & \multicolumn{1}{c}{\cellcolor[rgb]{.851,.851,.851}1.14E+03} & \multicolumn{1}{c}{\cellcolor[rgb]{.851,.851,.851}5.03E+04} & \cellcolor[rgb]{.851,.851,.851}5.14E+04 (34.0\%) \\
\cmidrule{1-6}\cmidrule{8-12}    TP9   & \multicolumn{1}{{c}}{\cellcolor[rgb]{.851,.851,.851}1.00E-06 ($\approx$)} & \multicolumn{1}{{c}}{\cellcolor[rgb]{.851,.851,.851}1.00E-06 ($\approx$)} & \multicolumn{1}{{c}}{1.03E+03 (+)} & \multicolumn{1}{{c}}{5.83E+04 (+)} & \multicolumn{1}{{c}}{5.93E+04 (+)} &       &\cellcolor[rgb]{.851,.851,.851} 1.00E-06 &\cellcolor[rgb]{.851,.851,.851} 1.00E-06 & \multicolumn{1}{c}{\cellcolor[rgb]{.851,.851,.851}5.93E+02} & \multicolumn{1}{c}{\cellcolor[rgb]{.851,.851,.851}3.45E+04} & \cellcolor[rgb]{.851,.851,.851}3.51E+04 (40.8\%) \\
\cmidrule{1-6}\cmidrule{8-12}    TP10  & \multicolumn{1}{{c}}{\cellcolor[rgb]{.851,.851,.851}1.00E-06 ($\approx$)} & \multicolumn{1}{{c}}{\cellcolor[rgb]{.851,.851,.851}1.00E-06 ($\approx$)} & \multicolumn{1}{{c}}{2.25E+03 (+)} & \multicolumn{1}{{c}}{1.06E+05 (+)} & \multicolumn{1}{{c}}{1.09E+05 (+)} &       &\cellcolor[rgb]{.851,.851,.851} 1.00E-06 &\cellcolor[rgb]{.851,.851,.851} 1.00E-06 & \multicolumn{1}{c}{\cellcolor[rgb]{.851,.851,.851}1.22E+03} & \multicolumn{1}{c}{\cellcolor[rgb]{.851,.851,.851}6.06E+04} & \cellcolor[rgb]{.851,.851,.851}6.18E+04 (43.1\%) \\
\cmidrule{1-6}\cmidrule{8-12}    +/$\approx$/- & 2/10/0 & 2/10/0 & 11/1/0 & 11/1/0 & 11/1/0 &       &       &       & \multicolumn{2}{r}{Average $R_{rs}$} & 28.8\% \\
    \bottomrule
\end{tabular}%
}
\end{table*}%

\begin{figure*}[htbp]
\centering
\includegraphics[width=\textwidth]{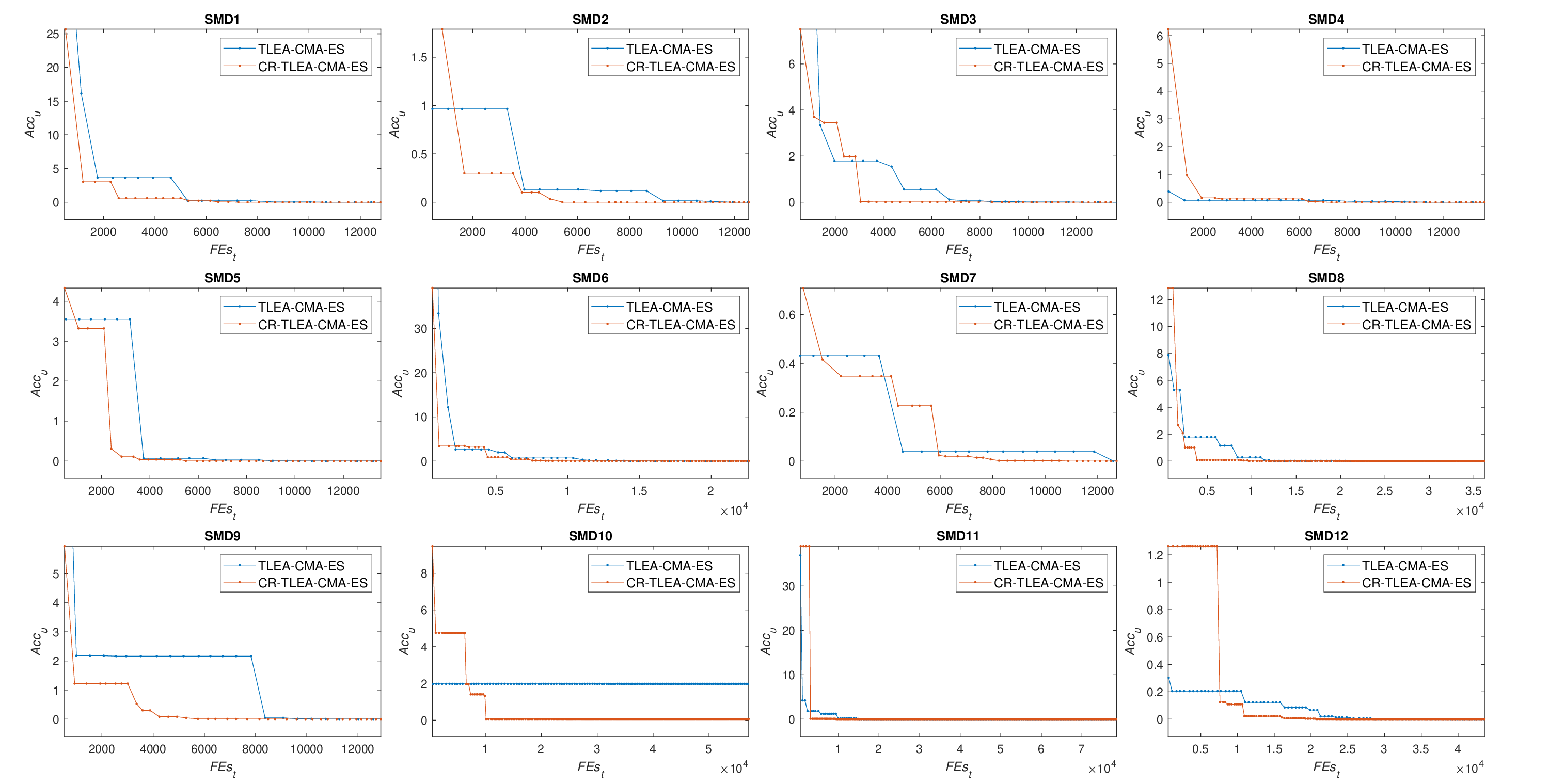}
\caption{Convergence curve of TLEA-CMA-ES and CR-TLEA-CMA-ES on SMD problems. For ease of observation, only the results within the termination $FEs_t$ of the faster-converging algorithm are presented.}
\label{fig:Convergence curve}
\end{figure*}

The comparison between CR-TLEA-CMA-ES and TLEA-CMA-ES on TP problems is presented in Table \ref{tab:TP_TLEACMAES_vs_CRTLEACMAES} as a representative case. 
TLEA-CMA-ES is selected as a typical base algorithm to illustrate the effectiveness of CR-BLEA, as similar trends were observed across other variants.
The same representative strategy is adopted in subsequent experiments to avoid redundancy and maintain clarity.
The results indicate that the CR-BLEA framework is also applicable to TP problems, as the algorithm achieves similar results to the original algorithm on most TP problems while significantly reducing the number of FEs.
In fact, since CR-BLEA does not interfere with the optimization strategy of the original algorithm, it maintains general applicability across different problems.

\subsection{Discussions}
The convergence curves of upper-level accuracy $Acc_u$ with respect to the total function evaluations $FEs_t$ during the optimization process of TLEA-CMA-ES and CR-TLEA-CMA-ES on SMD are presented in Fig. \ref{fig:Convergence curve}.
Consistent with the results in Table \ref{tab:TLEACMAES_vs_CRTLEACMAES}, the distance between the upper-level elitists and the actual optimal solution in CR-TLEA-CMA-ES decreases more rapidly, and the algorithm converges faster to the termination accuracy range.

In BLEAs, before evaluating a set of $x_u$ to update the upper-level accuracy, a series of lower-level iterations involving $FEs_l$ is required to find the corresponding $x_l^*$ for each $x_u$.
Therefore, the interval between two data points in $FEs_t$ is non-fixed.
Notably, the $FEs_t$ intervals between data points in CR-TLEA-CMA-ES are significantly smaller than those in TLEA-CMA-ES.
\begin{figure}[!htbp]
\centering
\includegraphics[width=0.7\linewidth]{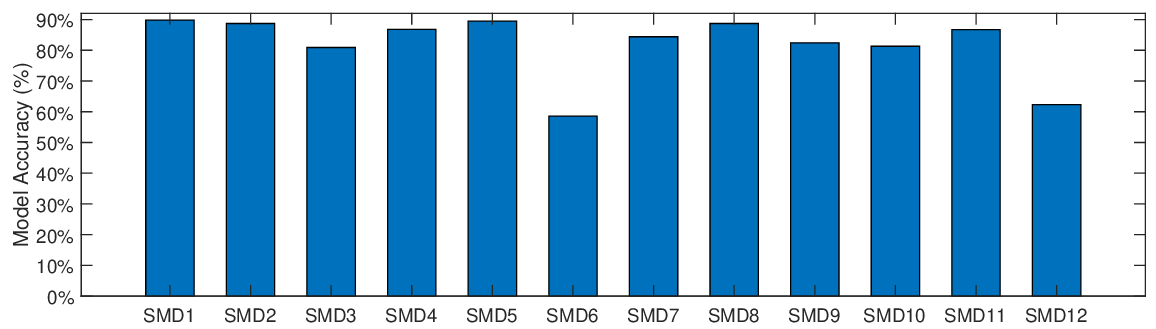}
\caption{Average Model Accuracy of CR-TLEA-CMA-ES on Different SMD Problems.}
\label{fig:Model accuracy}
\end{figure}
This difference stems from the strategy that, when handling a set of solutions $x_u$ of the same size, CR-TLEA-CMA-ES employs the trained contrastive network to identify top-ranked promising individuals, and allocates computing resources embodied as $FEs_l$ only to the lower-level tasks generated by these individuals, thus reducing the resource waste on unpromising tasks compared to TLEA-CMA-ES.

Since the contrastive ranking network in CR-BLEA is continuously updated with paired data generated from new populations, the training data used for the new model can serve as test data for the previous one.
Consequently, the average accuracy of all historical models throughout the optimization process can be used to assess the overall performance of the contrastive ranking model.

The average accuracy of CR-TLEA-CMA-ES on SMD problems is presented in Fig. \ref{fig:Model accuracy}.
On most problems, the average test accuracy of the model exceeds 80\% and approaches 90\% on SMD1 and SMD5.
On SMD6 and SMD12, the average accuracy of the model is only 58.6\% and 62.3\%, respectively, which confirms to the relatively lower average $FEs_t$ reduction reported in Table \ref{tab:TLEACMAES_vs_CRTLEACMAES}.
In SMD12, each lower level problem has multiple global optima, while SMD6 is even more challenging as it contains infinitely many global solutions at the lower level for any given upper-level vector.
The existence of multiple lower-level optima confuses the contrastive algorithm, as the $\psi(x_u)$ is multimodal and the relationship between the training pairs cannot be clearly defined.

The accuracy results indicate that the contrastive ranking model effectively extracts solution features and their relationships,
thereby effectively providing guidance for the ranking of newly generated solutions, which accordingly supports resource allocation.

\subsection{Ablation Study}
To investigate the effectiveness of the proposed contrastive ranking network and resampling strategy in CR-BLEA, two variants of CR-TLEA-CMA-ES are considered for ablation experiments:

$\bullet$ CR-TLEA-CMA-ES-v1: The contrastive ranking network is deactivated.
Instead of being selected by the model, tasks are randomly chosen from each new upper-level population for lower-level execution.

$\bullet$ CR-TLEA-CMA-ES-v2: The resampling strategy is removed.

\begin{table*}[tbp]
  \centering
  \caption{Ablation comparison between CR-TLEA-CMA-ES and the variants regarding the median results of the accuracy and the number of FEs on SMD problems in 21 runs.}
  \label{tab:Ablation_TLEACMAES_vs_CRTLEACMAES}
\resizebox{\textwidth}{!}{    \begin{tabular}{cccccccccccccccccc}
    \toprule
    \multirow{2}[4]{*}{Probelm} & \multicolumn{5}{c}{CR-TLEA-CMA-ES}    &       & \multicolumn{5}{c}{CR-TLEA-CMA-ES-v1} &       & \multicolumn{5}{c}{CR-TLEA-CMA-ES-v2} \\
\cmidrule{2-6}\cmidrule{8-12}\cmidrule{14-18}          & $Acc_u$   & $Acc_l$  & $FEs_u$  & $FEs_l$  & $FEs_t$  &       & $Acc_u$   & $Acc_l$  & $FEs_u$  & $FEs_l$  & $FEs_t$  &       & $Acc_u$   & $Acc_l$  & $FEs_u$  & $FEs_l$  & $FEs_t$ \\
\cmidrule{1-6}\cmidrule{8-12}\cmidrule{14-18}    SMD1  & \multicolumn{1}{p{5em}}{\cellcolor[rgb]{.851,.851,.851}1.00E-06} & \multicolumn{1}{p{5em}}{\cellcolor[rgb]{.851,.851,.851}1.00E-06} & \multicolumn{1}{p{5em}}{\cellcolor[rgb]{.851,.851,.851}1.77E+02} & \multicolumn{1}{p{5em}}{\cellcolor[rgb]{.851,.851,.851}1.26E+04} & \multicolumn{1}{p{5em}}{\cellcolor[rgb]{.851,.851,.851}1.28E+04} &       & \cellcolor[rgb]{.851,.851,.851}1.00E-06 ($\approx$) & \cellcolor[rgb]{.851,.851,.851}1.00E-06 ($\approx$) & 2.83E+02 (+) & 1.90E+04 (+) & 1.93E+04 (+) &       & \cellcolor[rgb]{.851,.851,.851}1.00E-06 ($\approx$) & \cellcolor[rgb]{.851,.851,.851}1.00E-06 ($\approx$) & 1.92E+02 ($\approx$) & 1.27E+04 ($\approx$) & 1.29E+04 ($\approx$) \\
\cmidrule{1-6}\cmidrule{8-12}\cmidrule{14-18}    SMD2  & \multicolumn{1}{p{5em}}{\cellcolor[rgb]{.851,.851,.851}1.00E-06} & \multicolumn{1}{p{5em}}{\cellcolor[rgb]{.851,.851,.851}1.16E-06} & \multicolumn{1}{p{5em}}{\cellcolor[rgb]{.851,.851,.851}1.72E+02} & \multicolumn{1}{p{5em}}{\cellcolor[rgb]{.851,.851,.851}1.24E+04} & \multicolumn{1}{p{5em}}{\cellcolor[rgb]{.851,.851,.851}1.25E+04} &       & \cellcolor[rgb]{.851,.851,.851}1.00E-06 ($\approx$) & 2.23E-06 (+) & 3.10E+02 (+) & 1.99E+04 (+) & 2.02E+04 (+) &       & \cellcolor[rgb]{.851,.851,.851}1.00E-06 ($\approx$) & 1.21E-06 ($\approx$) & 1.95E+02 (+) & 1.29E+04 (+) & 1.31E+04 (+) \\
\cmidrule{1-6}\cmidrule{8-12}\cmidrule{14-18}    SMD3  & \multicolumn{1}{p{5em}}{\cellcolor[rgb]{.851,.851,.851}1.00E-06} & \multicolumn{1}{p{5em}}{\cellcolor[rgb]{.851,.851,.851}1.00E-06} & \multicolumn{1}{p{5em}}{\cellcolor[rgb]{.851,.851,.851}1.85E+02} & \multicolumn{1}{p{5em}}{\cellcolor[rgb]{.851,.851,.851}1.35E+04} & \multicolumn{1}{p{5em}}{\cellcolor[rgb]{.851,.851,.851}1.37E+04} &       & \cellcolor[rgb]{.851,.851,.851}1.00E-06 ($\approx$) & \cellcolor[rgb]{.851,.851,.851}1.00E-06 ($\approx$) & 3.09E+02 (+) & 2.05E+04 (+) & 2.08E+04 (+) &       & \cellcolor[rgb]{.851,.851,.851}1.00E-06 ($\approx$) & \cellcolor[rgb]{.851,.851,.851}1.00E-06 ($\approx$) & 2.18E+02 (+) & 1.39E+04 (+) & 1.42E+04 (+) \\
\cmidrule{1-6}\cmidrule{8-12}\cmidrule{14-18}    SMD4  & \multicolumn{1}{p{5em}}{\cellcolor[rgb]{.851,.851,.851}1.00E-06} & \multicolumn{1}{p{5em}}{\cellcolor[rgb]{.851,.851,.851}1.89E-06} & \multicolumn{1}{p{5em}}{\cellcolor[rgb]{.851,.851,.851}2.12E+02} & \multicolumn{1}{p{5em}}{\cellcolor[rgb]{.851,.851,.851}1.35E+04} & \multicolumn{1}{p{5em}}{\cellcolor[rgb]{.851,.851,.851}1.37E+04} &       & \cellcolor[rgb]{.851,.851,.851}1.00E-06 ($\approx$) & 2.67E-06 ($\approx$) & 3.05E+02 (+) & 1.99E+04 (+) & 2.02E+04 (+) &       & \cellcolor[rgb]{.851,.851,.851}1.00E-06 ($\approx$) & 4.03E-06 (+) & 2.16E+02 ($\approx$) & 1.46E+04 (+) & 1.48E+04 (+) \\
\cmidrule{1-6}\cmidrule{8-12}\cmidrule{14-18}    SMD5  & \multicolumn{1}{p{5em}}{\cellcolor[rgb]{.851,.851,.851}1.00E-06} & \multicolumn{1}{p{5em}}{\cellcolor[rgb]{.851,.851,.851}1.17E-06} & \multicolumn{1}{p{5em}}{\cellcolor[rgb]{.851,.851,.851}2.09E+02} & \multicolumn{1}{p{5em}}{1.33E+04} & \multicolumn{1}{p{5em}}{1.35E+04} &       & \cellcolor[rgb]{.851,.851,.851}1.00E-06 ($\approx$) & 2.66E-06 ($\approx$) & 3.39E+02 (+) & 1.95E+04 (+) & 1.99E+04 (+) &       & \cellcolor[rgb]{.851,.851,.851}1.00E-06 ($\approx$) & 1.27E-06 ($\approx$) & 2.18E+02 ($\approx$) & \cellcolor[rgb]{.851,.851,.851}1.31E+04 ($\approx$) & \cellcolor[rgb]{.851,.851,.851}1.33E+04 ($\approx$) \\
\cmidrule{1-6}\cmidrule{8-12}\cmidrule{14-18}    SMD6  & \multicolumn{1}{p{5em}}{\cellcolor[rgb]{.851,.851,.851}1.00E-06} & \multicolumn{1}{p{5em}}{\cellcolor[rgb]{.851,.851,.851}1.00E-06} & \multicolumn{1}{p{5em}}{\cellcolor[rgb]{.851,.851,.851}3.86E+02} & \multicolumn{1}{p{5em}}{\cellcolor[rgb]{.851,.851,.851}2.22E+04} & \multicolumn{1}{p{5em}}{\cellcolor[rgb]{.851,.851,.851}2.26E+04} &       & \cellcolor[rgb]{.851,.851,.851}1.00E-06 ($\approx$) & \cellcolor[rgb]{.851,.851,.851}1.00E-06 ($\approx$) & 6.57E+02 (+) & 3.35E+04 (+) & 3.41E+04 (+) &       & \cellcolor[rgb]{.851,.851,.851}1.00E-06 ($\approx$) & \cellcolor[rgb]{.851,.851,.851}1.00E-06 ($\approx$) & 4.25E+02 (+) & 2.27E+04 (+) & 2.31E+04 (+) \\
\cmidrule{1-6}\cmidrule{8-12}\cmidrule{14-18}    SMD7  & \multicolumn{1}{p{5em}}{\cellcolor[rgb]{.851,.851,.851}1.00E-06} & \multicolumn{1}{p{5em}}{\cellcolor[rgb]{.851,.851,.851}1.00E-06} & \multicolumn{1}{p{5em}}{\cellcolor[rgb]{.851,.851,.851}1.76E+02} & \multicolumn{1}{p{5em}}{\cellcolor[rgb]{.851,.851,.851}1.25E+04} & \multicolumn{1}{p{5em}}{\cellcolor[rgb]{.851,.851,.851}1.27E+04} &       & 9.82E-02 (+) & 2.44E+02 (+) & 3.88E+02 (+) & 2.37E+04 (+) & 2.41E+04 (+) &       & \cellcolor[rgb]{.851,.851,.851}1.00E-06 ($\approx$) & \cellcolor[rgb]{.851,.851,.851}1.00E-06 ($\approx$) & 1.89E+02 ($\approx$) & 1.37E+04 (+) & 1.39E+04 (+) \\
\cmidrule{1-6}\cmidrule{8-12}\cmidrule{14-18}    SMD8  & \multicolumn{1}{p{5em}}{\cellcolor[rgb]{.851,.851,.851}1.00E-06} & \multicolumn{1}{p{5em}}{\cellcolor[rgb]{.851,.851,.851}1.00E-06} & \multicolumn{1}{p{5em}}{6.85E+02} & \multicolumn{1}{p{5em}}{3.55E+04} & \multicolumn{1}{p{5em}}{3.62E+04} &       & 1.59E-02 (+) & 1.72E-03 (+) & \cellcolor[rgb]{.851,.851,.851}5.66E+02 (-) & \cellcolor[rgb]{.851,.851,.851}3.33E+04 ($\approx$) & \cellcolor[rgb]{.851,.851,.851}3.38E+04 ($\approx$) &       & \cellcolor[rgb]{.851,.851,.851}1.00E-06 ($\approx$) & \cellcolor[rgb]{.851,.851,.851}1.00E-06 ($\approx$) & 7.14E+02 (+) & 3.69E+04 (+) & 3.76E+04 (+) \\
\cmidrule{1-6}\cmidrule{8-12}\cmidrule{14-18}    SMD9  & \multicolumn{1}{p{5em}}{\cellcolor[rgb]{.851,.851,.851}1.00E-06} & \multicolumn{1}{p{5em}}{\cellcolor[rgb]{.851,.851,.851}1.00E-06} & \multicolumn{1}{p{5em}}{\cellcolor[rgb]{.851,.851,.851}1.90E+02} & \multicolumn{1}{p{5em}}{\cellcolor[rgb]{.851,.851,.851}1.27E+04} & \multicolumn{1}{p{5em}}{\cellcolor[rgb]{.851,.851,.851}1.29E+04} &       & \cellcolor[rgb]{.851,.851,.851}1.00E-06 ($\approx$) & \cellcolor[rgb]{.851,.851,.851}1.00E-06 ($\approx$) & 3.23E+02 (+) & 2.20E+04 (+) & 2.23E+04 (+) &       & \cellcolor[rgb]{.851,.851,.851}1.00E-06 ($\approx$) & 1.17E-06 ($\approx$) & 2.00E+02 ($\approx$) & 1.32E+04 (+) & 1.34E+04 (+) \\
\cmidrule{1-6}\cmidrule{8-12}\cmidrule{14-18}    SMD10 & \multicolumn{1}{p{5em}}{\cellcolor[rgb]{.851,.851,.851}8.28E-03} & \multicolumn{1}{p{5em}}{2.14E-03} & \multicolumn{1}{p{5em}}{1.18E+03} & \multicolumn{1}{p{5em}}{5.59E+04} & \multicolumn{1}{p{5em}}{5.71E+04} &       & 1.60E+01 (+) & \cellcolor[rgb]{.851,.851,.851}1.63E-05 ($\approx$) & 1.23E+03 (+) & 5.44E+04 ($\approx$) & 5.56E+04 ($\approx$) &       & 6.36E-01 (+) & 1.57E+01 (+) & \cellcolor[rgb]{.851,.851,.851}1.02E+03 ($\approx$) & \cellcolor[rgb]{.851,.851,.851}4.91E+04 (-) & \cellcolor[rgb]{.851,.851,.851}5.01E+04 (-) \\
\cmidrule{1-6}\cmidrule{8-12}\cmidrule{14-18}    SMD11 & \multicolumn{1}{p{5em}}{\cellcolor[rgb]{.851,.851,.851}1.00E-06} & \multicolumn{1}{p{5em}}{\cellcolor[rgb]{.851,.851,.851}8.61E-05} & \multicolumn{1}{p{5em}}{1.64E+03} & \multicolumn{1}{p{5em}}{\cellcolor[rgb]{.851,.851,.851}7.69E+04} & \multicolumn{1}{p{5em}}{\cellcolor[rgb]{.851,.851,.851}7.86E+04} &       & 2.56E-04 (+) & 2.98E-04 (+) & 2.32E+03 (+) & 1.04E+05 (+) & 1.07E+05 (+) &       & \cellcolor[rgb]{.851,.851,.851}1.00E-06 ($\approx$) & 4.55E-04 (+) & \cellcolor[rgb]{.851,.851,.851}1.62E+03 ($\approx$) & 8.13E+04 (+) & 8.29E+04 (+) \\
\cmidrule{1-6}\cmidrule{8-12}\cmidrule{14-18}    SMD12 & \multicolumn{1}{p{5em}}{\cellcolor[rgb]{.851,.851,.851}1.00E-06} & \multicolumn{1}{p{5em}}{\cellcolor[rgb]{.851,.851,.851}3.00E-06} & \multicolumn{1}{p{5em}}{8.73E+02} & \multicolumn{1}{p{5em}}{\cellcolor[rgb]{.851,.851,.851}4.27E+04} & \multicolumn{1}{p{5em}}{\cellcolor[rgb]{.851,.851,.851}4.36E+04} &       & 1.54E-05 (+) & 3.96E-06 ($\approx$) & 1.27E+03 (+) & 6.44E+04 (+) & 6.56E+04 (+) &       & \cellcolor[rgb]{.851,.851,.851}1.00E-06 ($\approx$) & 4.72E-06 ($\approx$) & \cellcolor[rgb]{.851,.851,.851}8.41E+02 ($\approx$) & 4.37E+04 (+) & 4.46E+04 (+) \\
\cmidrule{1-6}\cmidrule{8-12}\cmidrule{14-18}    +/$\approx$/- &       &       &       &       &       &       & 5/7/0 & 4/8/0 & 11/0/1 & 10/2/0 & 10/2/0 &       & 1/11/0 & 3/9/0 & 4/8/0 & 9/2/1 & 9/2/1 \\
    \bottomrule
    \end{tabular}%
}
\end{table*}%

The comparison between CR-TLEA-CMA-ES and the variants in terms of accuracy and the number of FEs on SMD problems is summarized in Table \ref{tab:Ablation_TLEACMAES_vs_CRTLEACMAES}.
The statistical results indicate a significant difference between CR-TLEA-CMA-ES and its variants.
As can be observed, CR-TLEA-CMA-ES-v1 is inferior to CR-TLEA-CMA-ES in both accuracy and the number of FEs.
The random selection without preference leads to accuracy deterioration on many problems, as some promising lower-level tasks are discarded.
Additionally, CR-TLEA-CMA-ES-v1 consumes more computational resources due to lower-level optimization being performed on many unpromising tasks.

CR-TLEA-CMA-ES achieves better results than CR-TLEA-CMA-ES-v2 mainly in the number of function evaluations consumed. 
When the quality of the generated offspring population is relatively poor, CR-BLEA equipped with a resampling strategy can estimate the quality of the population using reference-based ranking and trigger offspring regeneration as needed, thereby further improving the efficiency of resource utilization.

\subsection{Real-world BLOPs Study}

To assess the effectiveness of the proposed framework on real-world problems, we test two practical cases from the fields of economics and management, which are among the most frequently encountered areas for bilevel optimization problems.
The first problem arises from the gold mining industry in Finland. The government at the upper level aims to maximize tax revenue and minimize environmental damage by controlling the tax rate, while the mining company at the lower level seeks to maximize profits and maintain its reputation by regulating mining production.
The second problem involves a company where the CEO at the upper level aims to maximize overall profits and improve product quality by managing the resource allocation, while branch managers at the lower level focus on branch profits and employee satisfaction through operational decision-making.
The description of these problems is detailed in \citep{islam2017enhanced}, and the scalarization method for multiple objectives follows \citep{he2018evolutionary}.

\begin{table}[tbp]
  \centering
  \caption{Performance comparison between TLEA-CMA-ES and CR-TLEA-CMA-ES on two real-world BLOPs.}
  \label{table:GM_DM}%
  \resizebox{0.8\linewidth}{!}{
    \begin{tabular}{cccccc}
    \toprule
    Problem & \multicolumn{2}{c}{Gold Mining} &       & \multicolumn{2}{c}{Decision Making} \\
\cmidrule{1-3}\cmidrule{5-6}    
    Algorithm & TLEA-CMA-ES & CR-TLEA-CMA-ES &       & TLEA-CMA-ES & CR-TLEA-CMA-ES \\
\cmidrule{1-3}\cmidrule{5-6}    
    $F$  & \cellcolor[rgb]{.851,.851,.851}-3.14E+02 ($\approx$) & \cellcolor[rgb]{.851,.851,.851}-3.14E+02 &       & \cellcolor[rgb]{.851,.851,.851}-1.09E+03 ($\approx$) & -1.08E+03 \\
    $f$  & 1.32E+03 (+) & \cellcolor[rgb]{.851,.851,.851}1.26E+03 &       & 1.03E+04 (+) & \cellcolor[rgb]{.851,.851,.851}6.90E+03 \\
    $FEs_u$  & 1.50E+03 (+) & \cellcolor[rgb]{.851,.851,.851}1.54E+02 &       & 2.51E+03 (+) & \cellcolor[rgb]{.851,.851,.851}8.82E+02 \\
    $FEs_l$  & 3.22E+04 (+) & \cellcolor[rgb]{.851,.851,.851}3.73E+03 &       & 1.12E+05 (+) & \cellcolor[rgb]{.851,.851,.851}5.77E+04 \\
    $FEs_t$ ($R_{rs}$) & 3.37E+04 (+) & \cellcolor[rgb]{.851,.851,.851}3.88E+03 (88.5\%) &       & 1.14E+05 (+) & \cellcolor[rgb]{.851,.851,.851}5.86E+04 (48.6\%) \\
    \bottomrule
    \end{tabular}%
}
\end{table}%

The comparison between TLEA-CMA-ES and CR-TLEA-CMA-ES on the two real-world problems is summarized in Table \ref{table:GM_DM}.
It can be observed that CR-TLEA-CMA-ES achieves similar upper-level results and better lower-level results compared to TLEA-CMA-ES, while significantly reducing the number of function evaluations at both levels.
As a result, total computational resource consumption is reduced by 88.5\% and 48.6\% in the gold mining problem and decision making problem, respectively.

\section{Conclusion}
\label{sec:Conclusion}
To reduce substantial resource consumption in bilevel evolutionary optimization, this paper introduces a knowledge-driven resource allocation framework named CR-BLEA, which aims to eliminate redundant lower-level optimizations.
A contrastive ranking network is developed to learn the relational patterns between paired solutions. 
By employing reference-based ranking, the model effectively identifies promising lower-level tasks, which are the only ones to be allocated resources for lower-level optimization.
Moreover, the reference-based ranking is capable of estimating the quality of new populations during the optimization process, which drives the algorithm to trigger offspring regeneration, if necessary, and thereby further enhances the efficiency of resource utilization.

The performance of CR-BLEA was tested on two test suites and two real-world problems by integrating five advanced BLEAs (i.e., TLEA-CMA-ES, TLEA-DE, BL-CMA-ES, MOTEA and BOC) into the proposed framework.
The results demonstrate that CR-BLEA can significantly reduce resource consumption while maintaining or even improving accuracy compared to the original algorithms.
Convergence trajectories, model performance and ablation experiments were investigated to analyze the causes for the performance improvement of the proposed framework.
For future work, we will work to introduce generative machine learning models into bilevel evolutionary optimization to facilitate convergence.
Additionally, extending the resource allocation mechanism to solve many-level optimization problems is another direction for future research.

\section*{CRediT authorship contribution statement}
\textbf{Dejun xu:} Conceptualization, Methodology, Software, Data curation, Writing - original draft. \textbf{Jijia Chen:} Methodology, Software, Data curation, Writing - review \& editing. \textbf{Gary G. Yen:} Supervision, Conceptualization, Writing - review \& editing. \textbf{Min Jiang:} Supervision, Methodology, Writing - review \& editing.

\section*{Declaration of competing interest}
The authors declare that they have no known competing financial interests or personal relationships that could have appeared to influence the work reported in this paper.

\section*{Acknowledgements}
This work was supported by the National Natural Science Foundation of China [Grant No.62276222]; and by the Public Technology Service Platform Project of Xiamen City [Grant No.3502Z20231043].

\bibliographystyle{elsarticle-num} 
\bibliography{cas-refs}

\end{document}